\documentclass[onecolumn]{IEEEtran}

\usepackage{xcolor}

\usepackage{amsopn}

\usepackage{lmodern}
\usepackage{amsmath}

\usepackage{graphicx}

\usepackage{amssymb,amsmath}

\usepackage{relsize}

\usepackage{upgreek}

\usepackage{subfigure}

\usepackage{url}

\usepackage{graphicx}

\usepackage{multicol}

\usepackage{algorithmic}
\usepackage{algorithm2e}

\pdfoutput=1
\usepackage{amsmath, amssymb}
\usepackage{graphicx,psfrag,epsf}
\usepackage{enumerate}
\usepackage[square,numbers]{natbib}
\usepackage{url} 

\newcommand{\argmax}{\mathop{\rm argmax}}

\usepackage{xcolor}

\usepackage{amsopn}

\usepackage{lmodern}
\usepackage{amsmath}

\usepackage{graphicx}

\usepackage{amssymb,amsmath}

\usepackage{relsize}

\usepackage{upgreek}

\usepackage{subfigure}

\usepackage{url}

\usepackage{graphicx}

\usepackage{algorithmic}
\usepackage{algorithm2e}

\pdfoutput=1
\usepackage{amsmath, amssymb}
\usepackage{graphicx,psfrag,epsf}
\usepackage{enumerate}
\usepackage{natbib}
\usepackage{url} 

\def\C{\mathcal{C}}

\def\sample{\{\mathbf{x}_i\}_{i\in [n]}}

\def\R{\mathbb{R}}

\def\x{\mathbf{x}}

\newcommand{\nk}[1]{\mathcal{N}_k(#1)}

\ifCLASSINFOpdf
\else
\fi

\usepackage{multicol}
\usepackage{lipsum}

\newcommand\blfootnote[1]{%
  \begingroup
  \renewcommand\thefootnote{}\footnote{#1}%
  \addtocounter{footnote}{-1}%
  \endgroup
}

\begin{document}
%
\title{Nearest Neighbour Equilibrium Clustering}
%
%
%

\author{
David~P.~Hofmeyr
\thanks{D.P. Hofmeyr was with the School of Mathematical Sciences, Lancaster University, United Kingdom.\\
\email{d.p.hofmeyr@lancaster.ac.uk}\\
\\
 \textit{\textcopyright 2025 IEEE. Personal use of this material is permitted. Permission from IEEE must be obtained for all other uses, in any current or future media, including reprinting/republishing this material for advertising or promotional purposes, creating new collective works, for resale or redistribution to servers or lists, or reuse of any copyrighted component of this work in other works.}}
}

\begin{center}
{\Huge \bf Nearest Neighbour Equilibrium Clustering}
\end{center}
\vspace{20pt}
{\bf David P. Hofmeyr\\
 School of Mathematical Sciences, Lancaster University, UK}
\vspace{20pt}

\begin{multicols}{2}

\begin{abstract}
A novel and intuitive nearest neighbours based clustering algorithm is introduced, in which a cluster is defined in terms of an equilibrium condition which balances its size and cohesiveness. The formulation of the equilibrium condition allows for a quantification of the strength of alignment of each point to a cluster, with these cluster alignment strengths leading naturally to a model selection criterion which renders the proposed approach fully automatable. The algorithm is simple to implement and computationally efficient, and produces clustering solutions of extremely high quality in comparison with relevant benchmarks from the literature. {\tt R} code to implement the approach is available from \url{https://github.com/DavidHofmeyr/NNEC}.
\end{abstract}


\blfootnote{\textit{\textcopyright 2025 IEEE. Personal use of this material is permitted. Permission from IEEE must be obtained for all other uses, in any current or future media, including reprinting/republishing this material for advertising or promotional purposes, creating new collective works, for resale or redistribution to servers or lists, or reuse of any copyrighted component of this work in other works.}}

%

\medmuskip = 1.5mu

\section{Introduction}
\label{sec:intro}

Clustering, or cluster analysis, is the task of partitioning a set of data into groups, or \textit{clusters}, which are seen to be relatively more homogeneous than the data as a whole. Clustering is one of the fundamental data analytic tasks, and forms an integral component of exploratory data analysis. Clustering is also of arguably increasing relevance, as data are increasingly being collected/generated from automated processes, where typically very little prior knowledge is available, making exploratory methods a necessity.

In the classical clustering problem there is no explicit information about how the data should be grouped, and various interpretations of how clusters of points may be defined have led to the development of a very large number of methods for identifying them. Almost universally, however, clusters are determined from the geometric properties of the data, with pairs of points which are near to one another typically being seen as likely to be in the same cluster and pairs which are distant more likely to be in different clusters. With this interpretation, which is extremely intuitive, there still remains a lot of flexibility in how to determine entire clusters. Some popular classes of methods include (i) centroid-based clustering, in which clusters are defined as compact collections of points around a central prototype~\citep{leisch2006}; (ii) model-based clustering, in which clusters are aligned with the components of a mixture model, with each component typically having a relatively simple parametric structure~\citep{fraley2002model}; (iii) density-based clustering, in which clusters are defined as regions of high data density which are separated from other clusters by regions of relatively low density~\citep{campello2020density}; and (iv) graph-based and spectral clustering, in which clusters are defined as highly connected sub-graphs which are at most weakly connected to other clusters~\citep{Luxburg2007}.\\
\\
Clustering using the nearest neighbour relationships in a data set is intuitively pleasing, as they concisely capture the local geometric structure in the data. It is completely natural to have as one of the objectives of a clustering procedure to obtain a solution in which points are clustered together with a high proportion of their nearest neighbours. However, it should be apparent that whether or not this objective is achievable for all points simultaneously will be heavily data dependent. If clusters of points are very well separated then it should be possible for every point to be clustered with all of its nearest neighbours, provided the number of neighbours classified as among the ``nearest'' does not exceed the size of any of the clusters. On the other hand, if the density of points near any shared boundaries between clusters is high, then the nearest neighbours of points near these boundaries will most often overlap with all, or most of the clusters sharing the boundary, leading to cluster membership of these points being somewhat ambiguous. One might avoid this by considering only the single nearest neighbour of every point, since a singleton cannot overlap with multiple disjoint clusters. Clustering using only the single nearest neighbours is typically achieved with the Single Linkage Agglommerative clustering model~\citep{singleLinkage}, however, perhaps unsurprisingly, such an approach is extremely sensitive to noise and is generally not the preferred approach except in domains where there is knowledge that clusters can be characterised in this way. More recently clustering using the spectral properties of the nearest neighbour graph of the data has become popular, via the appropriately named Spectral Clustering~\citep[SC]{Luxburg2007,NNSC}. Indirectly SC using nearest neighbours seeks to minimise the number of pairs of near neighbours \textit{not} clustered together, while simultaneously ensuring each cluster has a high degree of internal cohesiveness, determined in terms of nearest neighbours being ``connected'' to one another. In the past few years a number of approaches have been developed which seek to resolve the issue of ambiguity of cluster membership near shared cluster boundaries by ``peeling away'' points which are likely to belong to these ``ambiguous regions'', before clustering the remaining more obviously clusterable points using a modified neighbour-based graph~\citep{BPC, SNNC}.\\
\\
In this paper we introduce a novel approach for clustering using nearest neighbours, in which clusters are characterised by an equilibrium condition which balances their size and cohesiveness. For a given number of neighbours and balancing threshold parameter the proposed approach automatically determines an appropriate number of clusters to extract from a data set. Moreover, we describe a natural criterion for selecting appropriate values for these two tuning parameters, making the proposed approach fully automatable. 
Our approach is intuitive, straightforward, and very simple to implement. To document its practical relevance, we evaluate its performance on a large collection of publicly available data sets which have been used frequently in the clustering literature. The results show remarkable promise, and the proposed approach achieves very high degrees of clustering accuracy in comparison with existing approaches, both classical and recent.\\
\\
The remainder of this paper is organised as follows. In Section~\ref{sec:method} we detail our formulation of \textit{equilibrium clusters}, and describe our methodology. We also provide pseudo-code giving a complete and explicit description of our algorithm used for performing clustering. We go on to introduce a simple criterion which can be used to automatically determine appropriate settings for our method's tuning parameters. In Section~\ref{sec:experiments} we document the strong clustering performance of the approach on a large collection of data sets in the public domain, and in comparison with numerous alternative approaches from the literature. Finally, we conclude with a brief discussion in Section~\ref{sec:conclusions}.

\section{Equilibrium Clusters} \label{sec:method}

In this section we give an overview of the proposed formulation, and describe a simple and intuitive approach for automatically selecting its tuning parameters. The method is based on the simple idea of finding clusters which satisfy an equilibrium condition, which states that all points in a cluster must also have at least a certain number of their neighbours in the cluster, with the threshold proportional to the size of the cluster. 
We use the term ``equilibrium'' since this characterisation of a cluster describes a balancing of the size of the cluster against its cohesiveness, with a cohesive cluster interpreted as having its points grouped along with a high proportion of their nearest neighbours. In addition, how we find these equilibrium clusters is through an iterative process in which the cluster is initially expanded from a seed point, and points which fit cohesively into the cluster are ``absorbed'', up to a point where it begins to stabilise and some points may move in and out as it reaches an appropriate size and level of cohesiveness. We make this more precise in the following.\\
\\
Let $\sample$ be a sample of points in a metric space, and for $j \in [n], k \in [n-1]$ let $\nk{\x_j}$ be the set of indices of the $k$ nearest points to $\x_j$ from among $\{\x_i\}_{i\in [n]\setminus\{j\}}$\footnote{ties in the distances between points may be broken arbitrarily}. Note that we have used the notation $[n]$ to denote the first $n$ natural numbers, i.e., $[n] = \{1, ..., n\}$. Then, for a given $\lambda > 0$, we say that a cluster with indices $\C \subset [n]$ is a $(\lambda, k)$ equilibrium cluster if
\begin{align}\label{eq:equilibrium}
    \frac{\left|\nk{\x_j} \cap \C\right|}{k} > \lambda \frac{|\C|}{n} \geq \frac{\left|\nk{\x_l} \cap \C\right|}{k} \ \forall j \in \C, l \in [n]\setminus \C,
\end{align}
where $|\cdot|$is the counting measure, i.e., $|\C|$ is the number of elements in $\C$. For brevity, when there is no ambiguity, we will simply use the term equilibrium cluster and suppress the dependence on $\lambda$ and $k$.\\
\\
Intuitively, points near the cores of clusters will typically have all of their nearest neighbours in the cluster, unless the value for $k$ is extremely large. Moreover, points well away from a cluster will generally have none, or very few of their neighbours in the cluster. It is points near the boundaries of clusters, therefore, for which the equilibrium condition described in Ineq.~(\ref{eq:equilibrium}) is critical. When clusters are well separated, then even the points on their boundaries will have all, or almost all of their neighbours falling within the same clusters. In these cases any sensible setting for $\lambda$ will correctly identify these clusters as equilibrium clusters. 
%
On the other hand, when clusters are less clearly defined then boundary points will only have a smaller fraction of their neighbours in the cluster. Perhaps more precisely, where the boundaries of clusters actually lie may be somewhat ambiguous, and in this case the setting of $\lambda$ strongly dictates how equilbrium clusters form around a core of ``obvious'' points. 
%
Clearly, therefore, an appropriate setting for $\lambda$ will generally not be known \textit{a priori}, and the appropriateness of different settings for $\lambda$ will also depend on the setting of $k$. We therefore describe a simple approach for selecting both of these parameters, in Section~\ref{sec:tuning}. In addition, as we describe in Section~\ref{sec:algorithm}, the number of clusters extracted is fully determined by settings of $\lambda$ and $k$, making our approach fully automatable.


\subsection{Finding Equilibrium Clusters}

As mentioned previously, we use an iterative procedure to identify equilibrium clusters. It is worth pointing out, however, that for any fixed $\lambda$ the collections of values which can be assumed by the terms $\frac{|\nk{\x_i}\cap \C|}{k}$ and $\lambda \frac{|\C|}{n}$ are both discrete, but are on different levels of granularity. As a result, the iterative approach we describe may not reach perfect equilibrium, since points may cycle in and out as repeated small changes to $\frac{|\C|}{n}$ across iterations cause these points to alternate between satisfying and not satisfying the condition in Eq.~(\ref{eq:equilibrium}). Typically, when this occurs, it applies only to a very small number of points, and we have not found stopping at an arbitrary point once this cycling nature is identified to be detrimental to the performance of the method. The reason for this is that these cycling points, as expected, typically arise near the boundaries of the cluster and are often more obvious members of at least one other cluster, meaning their final allocation is often the same regardless.

\begin{figure*}[t]
    \centering
    \subfigure[1 iteration]{\includegraphics[width=0.22\linewidth]{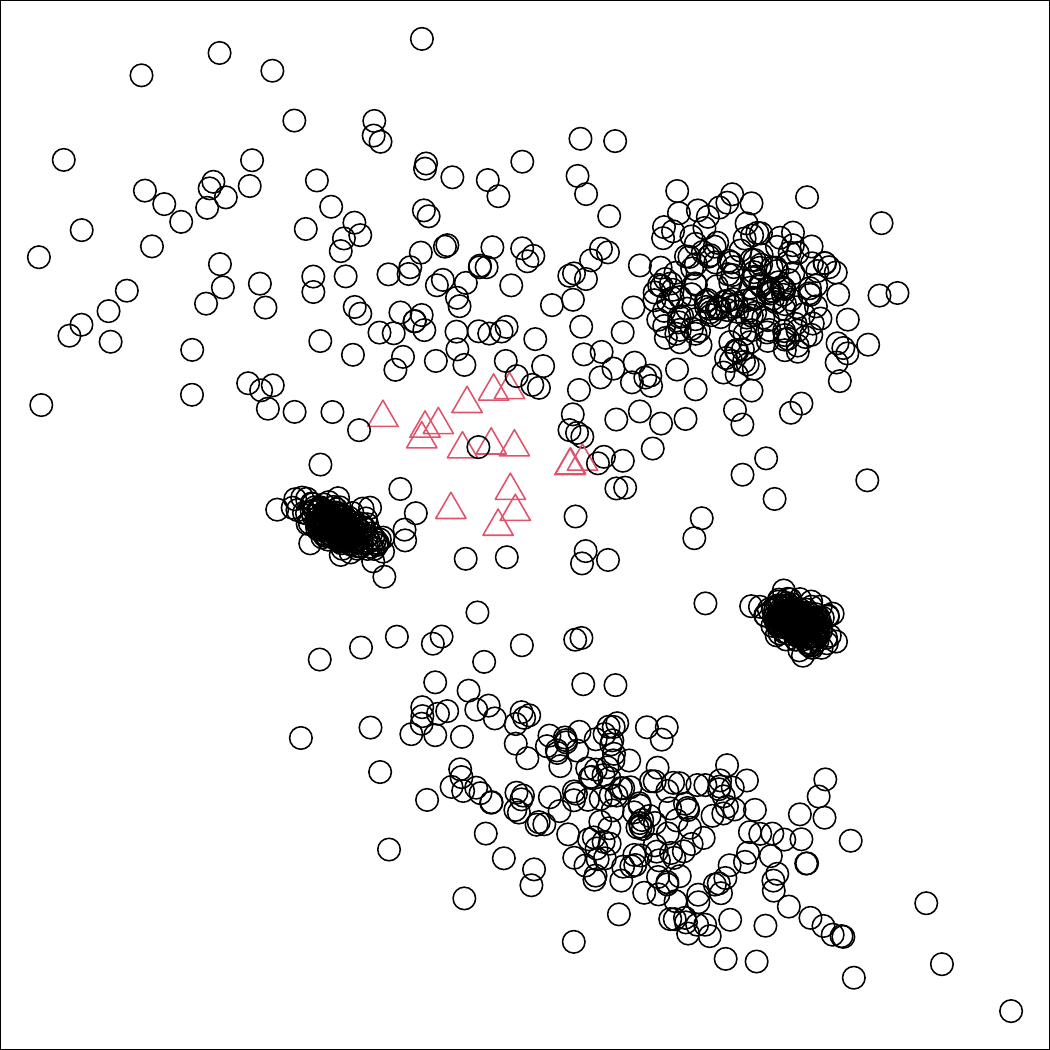}\label{fig:grow_cluster1}}
    \subfigure[2 iterations]{\includegraphics[width=0.22\linewidth]{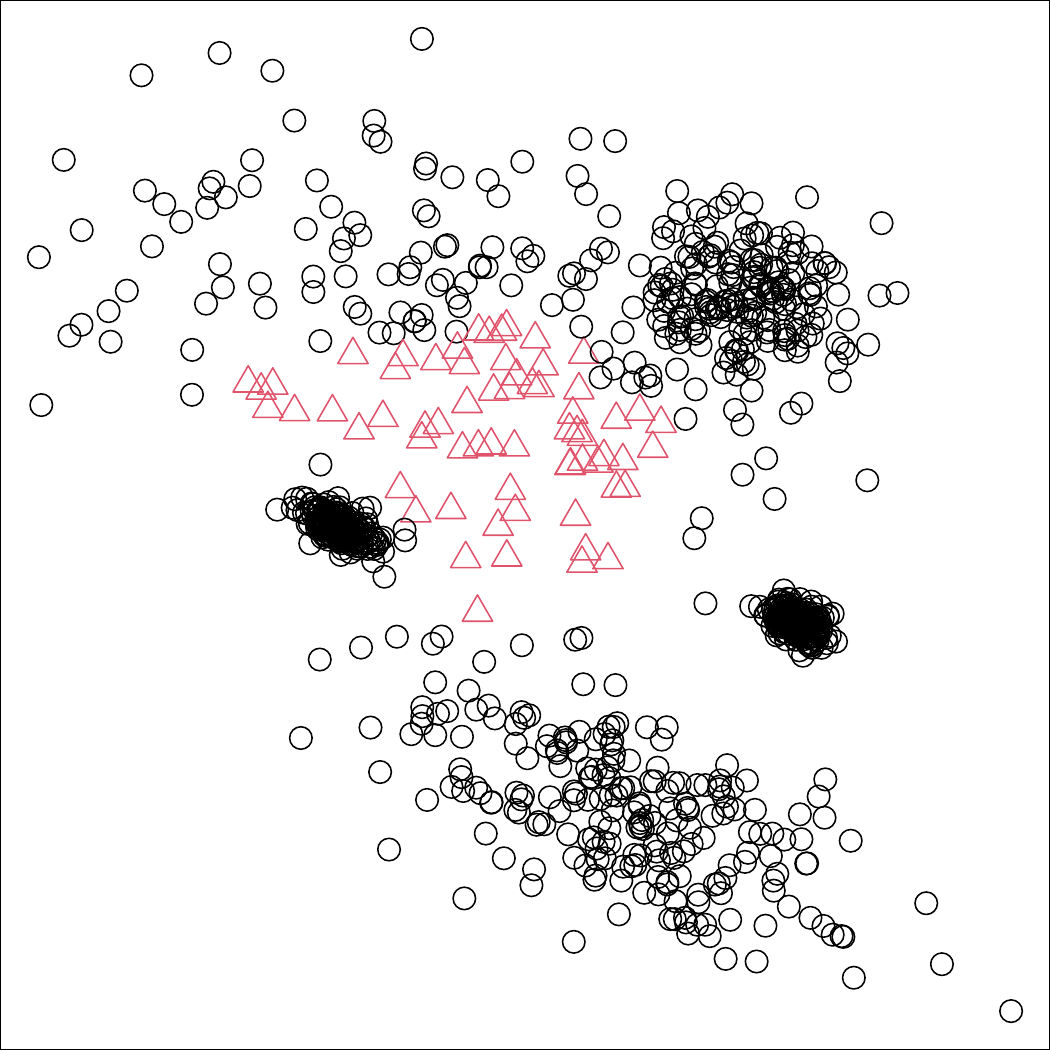}\label{fig:grow_cluster2}}
    \subfigure[5 iterations]{\includegraphics[width=0.22\linewidth]{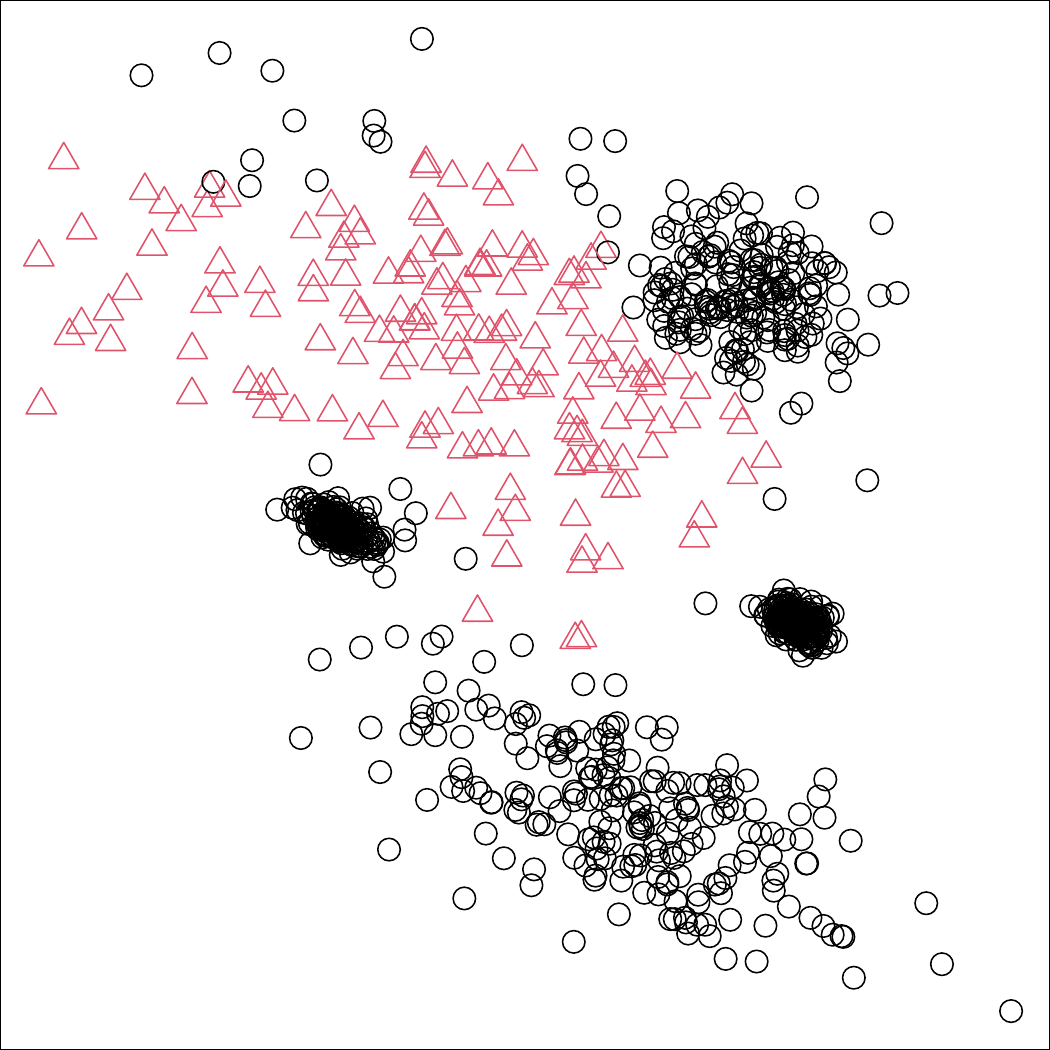}\label{fig:grow_cluster3}}
    \subfigure[9 iterations]{\includegraphics[width=0.22\linewidth]{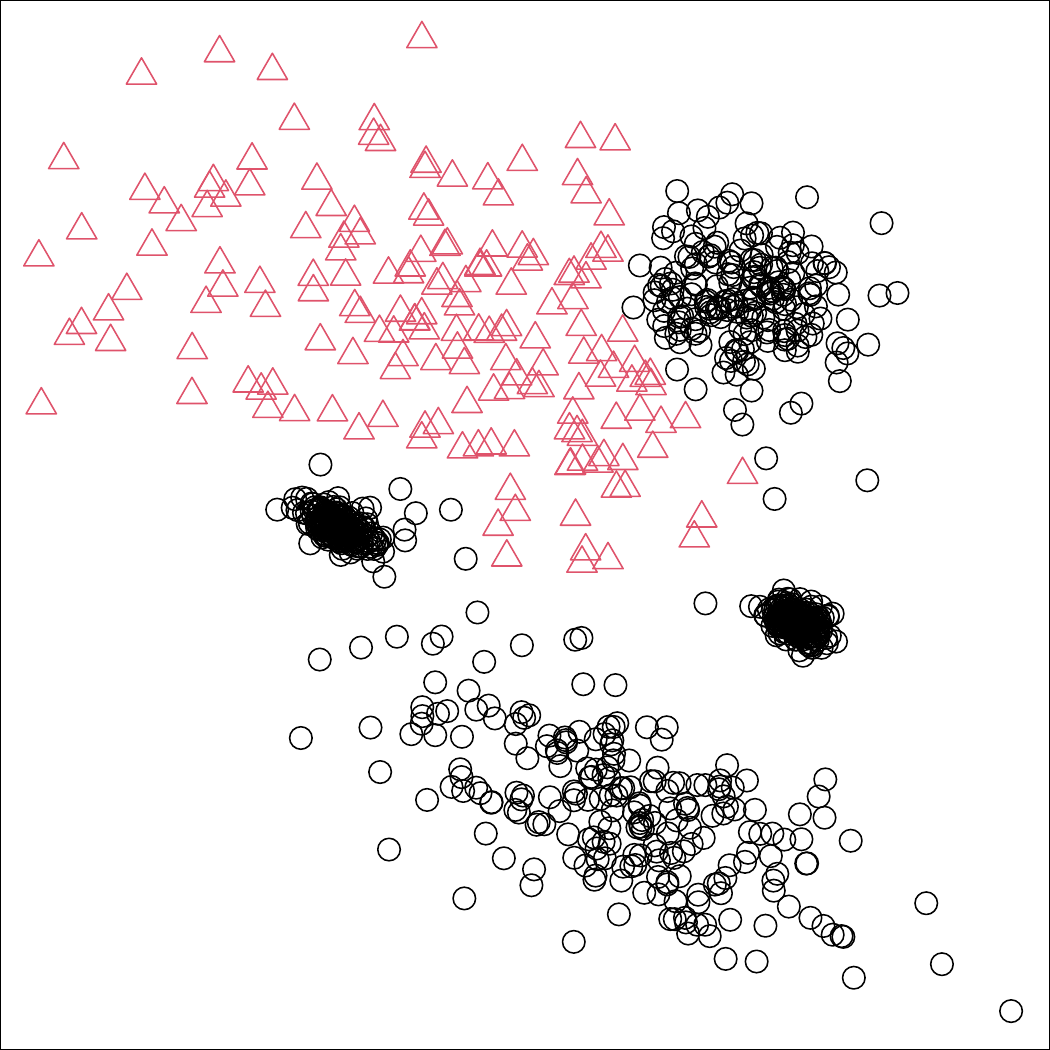}\label{fig:grow_cluster4}}
    \caption{Running of  Algorithm~\ref{alg:grow} for $k = 25$ and $\lambda = 2$, after different numbers of iterations}
    \label{fig:grow_cluster}
\end{figure*}

The iterative algorithm we use to obtain (or approximate, in the presence of cycling) equilibrium clusters is given in Algorithm~\ref{alg:grow}. In theory cycling over any number of iterations might be possible, which may be difficult (or computationally demanding) to identify. We therefore only check for cycling within the previous $r$ iterations, where in our implementations we set $r = 5$, however as a result we also impose a maximum number of iterations since if cycling over more than $r$ iterations is present the iteration described may never terminate.
\begin{algorithm*}
    INPUT: Sample of points ($\sample$); seed ($j \in [n]$); Number of neighbours ($k \in [n-1]$); threshold parameter ($\lambda > 0$); maximum detectable cycle length ($r \in \mathbb{N}$); maximum number of iterations ($T \in \mathbb{N}$)
    \begin{algorithmic}
        \STATE \textit{\#\#\# Initialise iteration counter and cluster}
        \STATE $t \gets 0$
        \STATE $\C^{(0)} \gets \{j\}$
        \STATE \textit{\#\#\# Iterate until equilibrium, cycling, or maximum number of iterations}
        \WHILE{$\C^{(t)} \not \in \{\C^{(t-i)}\}_{i \in [r]}$ and $t \leq T$}
        \STATE $t \gets t + 1$
        \STATE $\C^{(t)} \gets \left\{ i \in [n] \big| \frac{|\nk{\x_i}\cap \C^{(t-1)}|}{k} > \lambda \frac{|\C^{(t-1)}|}{n}\right\}$
        \ENDWHILE
        \RETURN $\C^{(t)}$
    \end{algorithmic}
    \caption{EquilibriumCluster algorithm}\label{alg:grow}
\end{algorithm*}
Figure~\ref{fig:grow_cluster} shows a few stages of the application of this algorithm to a simple two-dimensional data set, containing five clusters corresponding to the components of a Gaussian mixture model in which the components have varying scale. The seed point lies near the middle of the figure, and the settings of $k$ and $\lambda$ are 25 and 2, respectively. Note that these values were selected automatically using the approach we describe in Section~\ref{sec:tuning}. After a single iteration (Figure~\ref{fig:grow_cluster1}) a small cluster of points, shown with \textcolor{red}{$\triangle$}'s, has emerged. As the number of iterations increases, more points are added, and after 9 iterations (Figure~\ref{fig:grow_cluster4}) the algorithm has converged, and the equilibrium cluster agrees with what we may likely have identified to be a cluster by eye. 

\subsection{A Complete Clustering Algorithm}\label{sec:algorithm}

In the previous subsection we described a simple iterative algorithm for finding (or approximating) individual equilibrium clusters. However it should be clear that the cluster will depend crucially on its seed, and moreover multiple different seeds could lead to the same clusters being re-discovered. Decisions therefore need to be made regarding how to select seeds to reduce this risk, and also how many clusters to extract. In addition, note that the characterisation of an equilibrium cluster does not imply uniqueness of membership, and in fact typically points near the boundaries of equilibrium clusters lie in multiple such clusters. A decision for how to produce a final (flat) clustering solution, from a collection of equilibrium clusters, therefore needs to be made.\\
\\
It is natural to choose a minimal number of equilibrium clusters needed to ensure all points are allocated to at least one. Although we cannot guarantee global minimality, due to the large number of equilibrium clusters which could arise from all potential seeds, we follow this principle in iteratively adding new equilibrium clusters until each point is included in at least one.
Further aligning with the objective of finding a minimal set of equilibrium clusters to cover $\sample$, it is natural to choose seeds which are likely to lead to the largest clusters. Note that after one iteration of Algorithm~\ref{alg:grow} (assuming $\lambda$ is not set so large that no equilibrium clusters can exist) we will have $\C^{(1)} = \{i \in [n] | j \in \nk{\x_i}\}$, where as in the description of the algorithm we have used $j$ to denote the seed of the cluster. That is, the cluster contains all points which have the seed point among their $k$ nearest neighbours. It is intuitive, therefore, to select seeds which have a large number of such points, as these lead initially to the most rapid growth of the cluster.
Finally, although occurring only rarely, we have observed situations in which an equilibrium cluster arising from Algorithm~\ref{alg:grow} does not contain its seed. If this is the case, then using this point as a seed again will result in the re-discovery of the same equilibrium cluster. To avoid this, when a cluster does not contain its seed, we also add a singleton cluster which contains only that seed point.\\
%
%
\\
Now, once a collection of equilibrium clusters has been obtained which covers the entire sample, a decision must be made how to treat their overlap. Fortunately, the criterion characterising an equilibrium cluster, in Ineq.~(\ref{eq:equilibrium}), provides a natural way to measure the strength of a point's membership to the cluster. Specifically, the strength of membership of a point, $\x_i$, to equilibrium cluster $\C$ may be defined by
\begin{align*}
    s_{i,\C}:= \left(\frac{|\nk{\x_i}\cap \C|}{k} - \lambda \frac{|\C|}{n}\right)_+,
\end{align*}
where the subscript ``$+$'' indicates the positive part of a real number, i.e., $(x)_+ = \max\{0, x\}; x \in \R$. Note that this is defined even for points which do not belong to the cluster, since if a point does not satisfy the condition for membership then the quantity in the bracket above is non-positive, and hence its strength of membership is zero. A final allocation of points to clusters, via their maximum membership, is therefore not only possible but entirely natural.\\
\\
%
%
A complete clustering algorithm based on the principles given above is outlined in Algorithm~\ref{alg:cluster_all}. The algorithm uses Algorithm~\ref{alg:grow} as a subroutine, indicated by the notation EquilibriumCluster$(\cdot)$. We have also used the notation $\overline{A}$ to indicate the complement of a set $A$. In practice, for the final assignment we break any ties in the membership strengths arbitrarily, using the smallest index. Generally this affects very few points, and more sophisticated approaches like using the membership strengths of their neighbours did not appreciably affect the performance of the algorithm.

\begin{figure*}[t]
    \centering
    \subfigure[]{\includegraphics[width=0.15\linewidth]{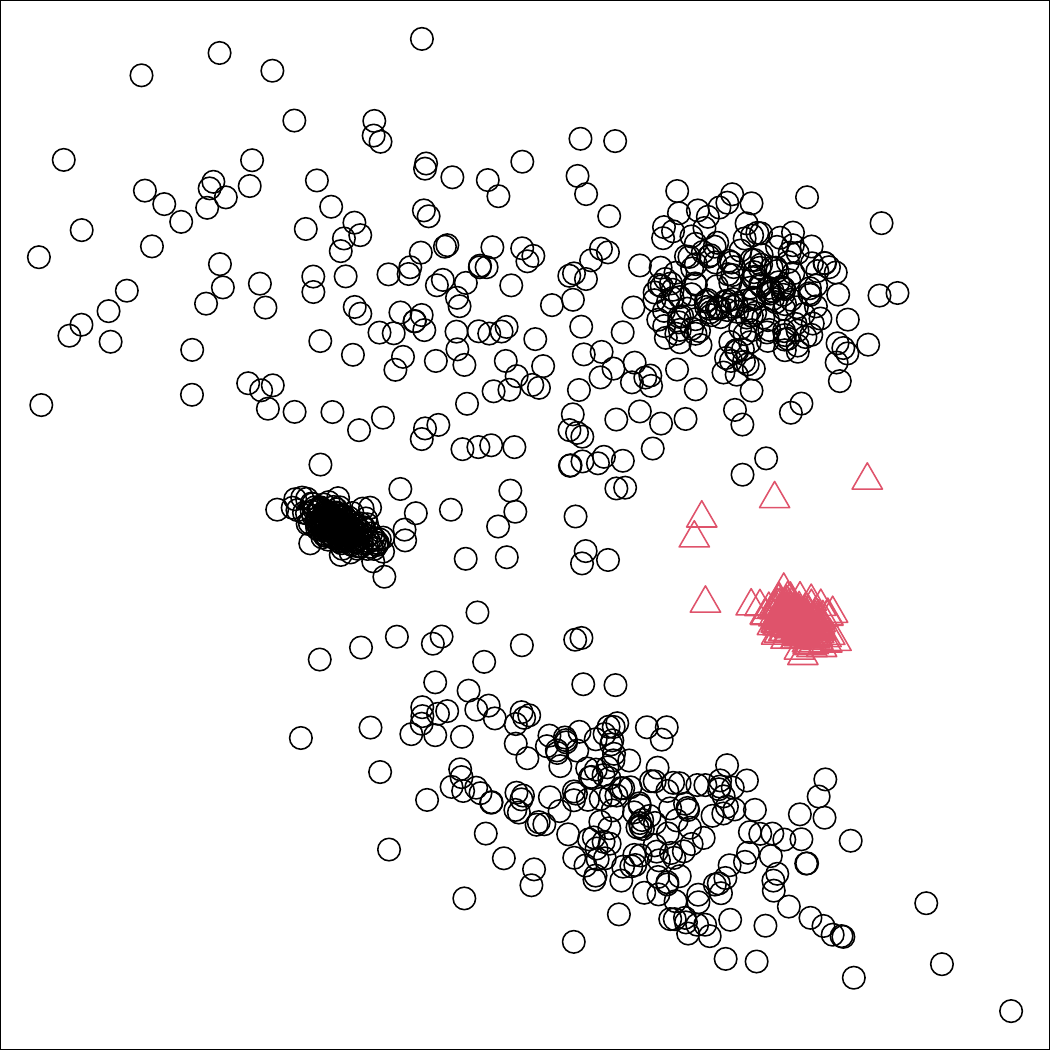}}
    \subfigure[]{\includegraphics[width=0.15\linewidth]{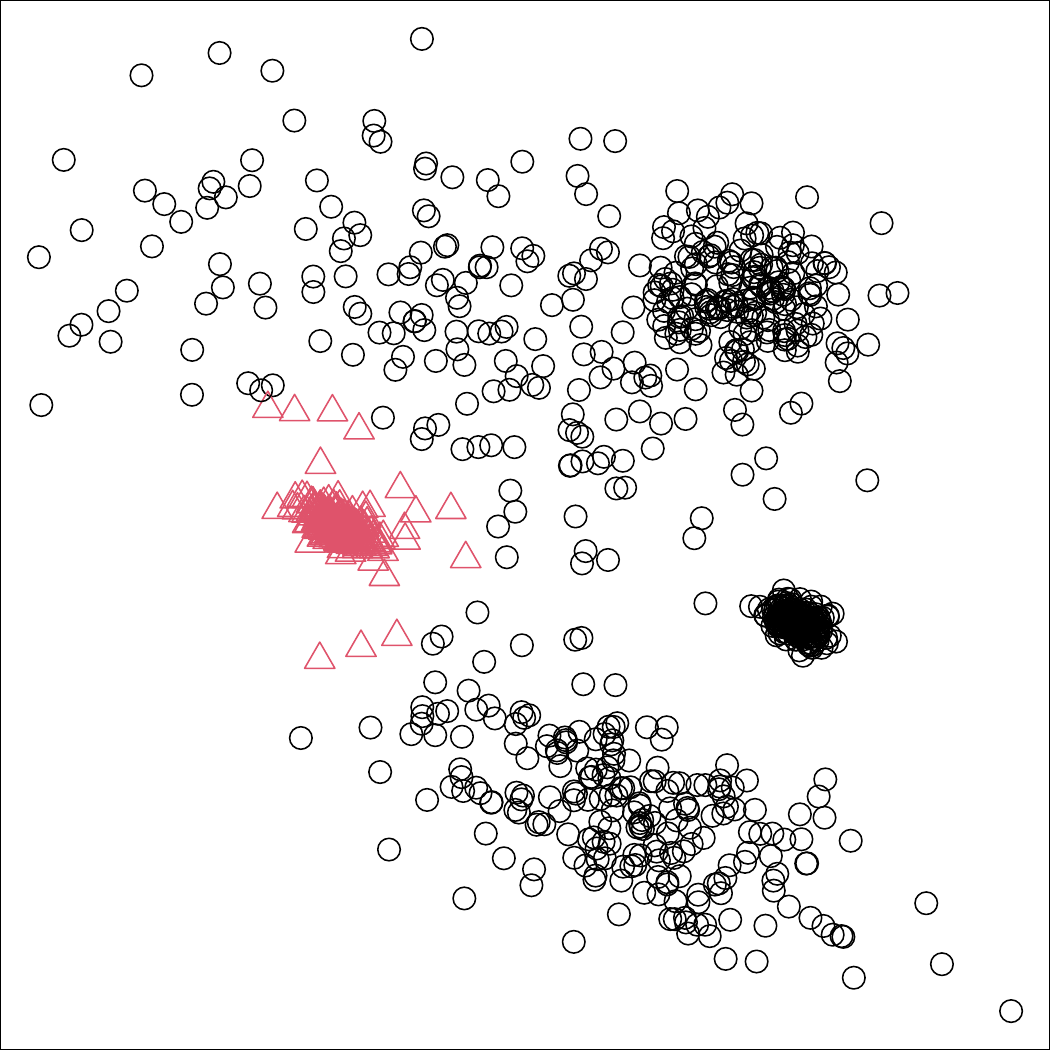}}
    \subfigure[]{\includegraphics[width=0.15\linewidth]{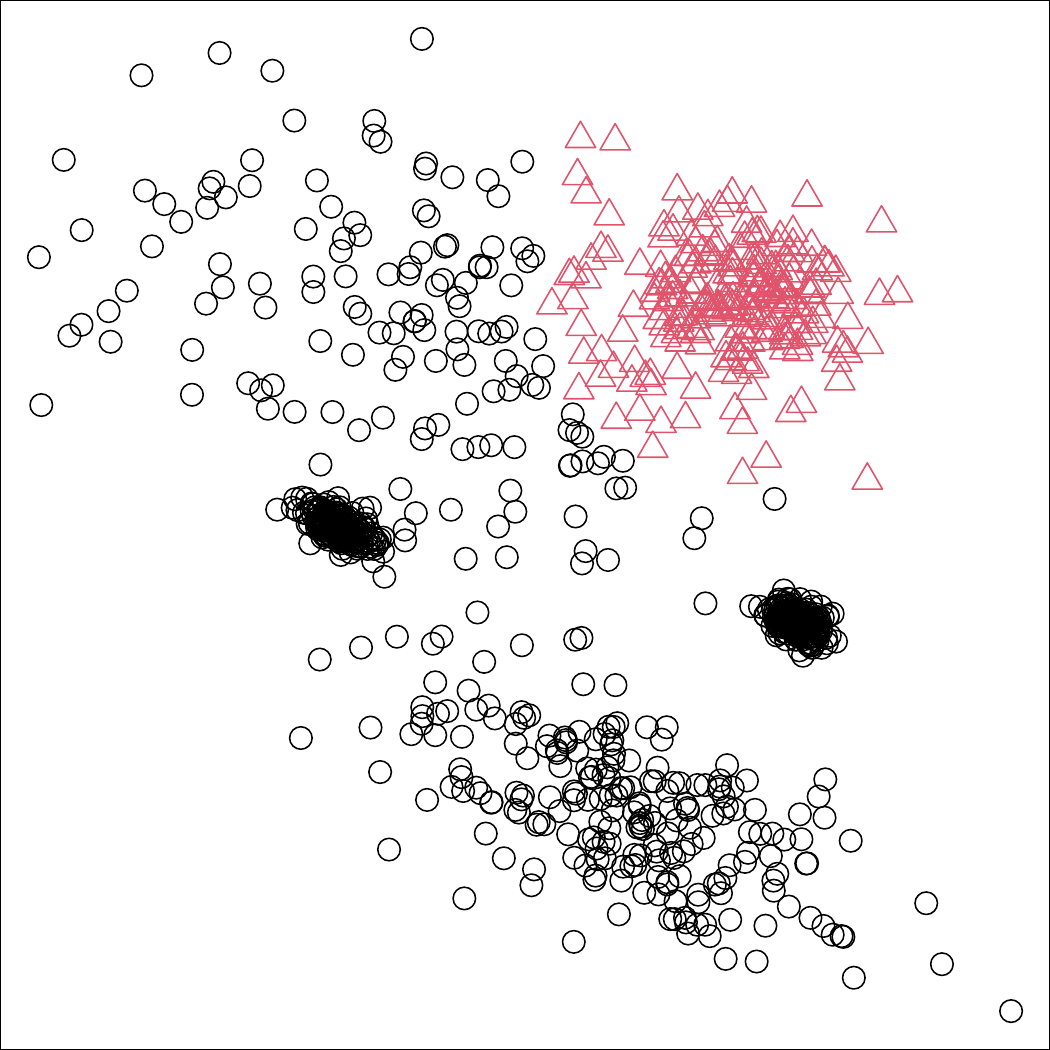}}
    \subfigure[]{\includegraphics[width=0.15\linewidth]{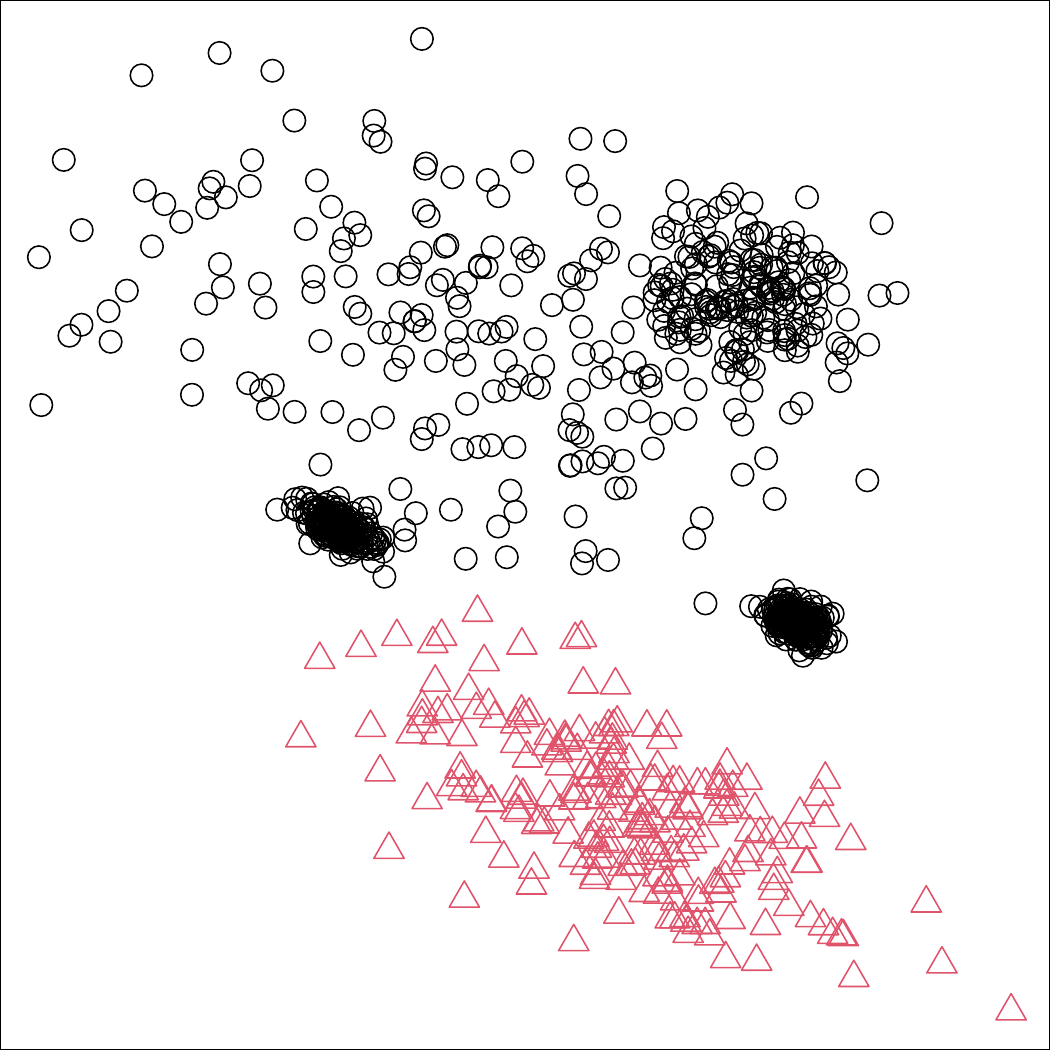}}
    \subfigure[]{\includegraphics[width=0.15\linewidth]{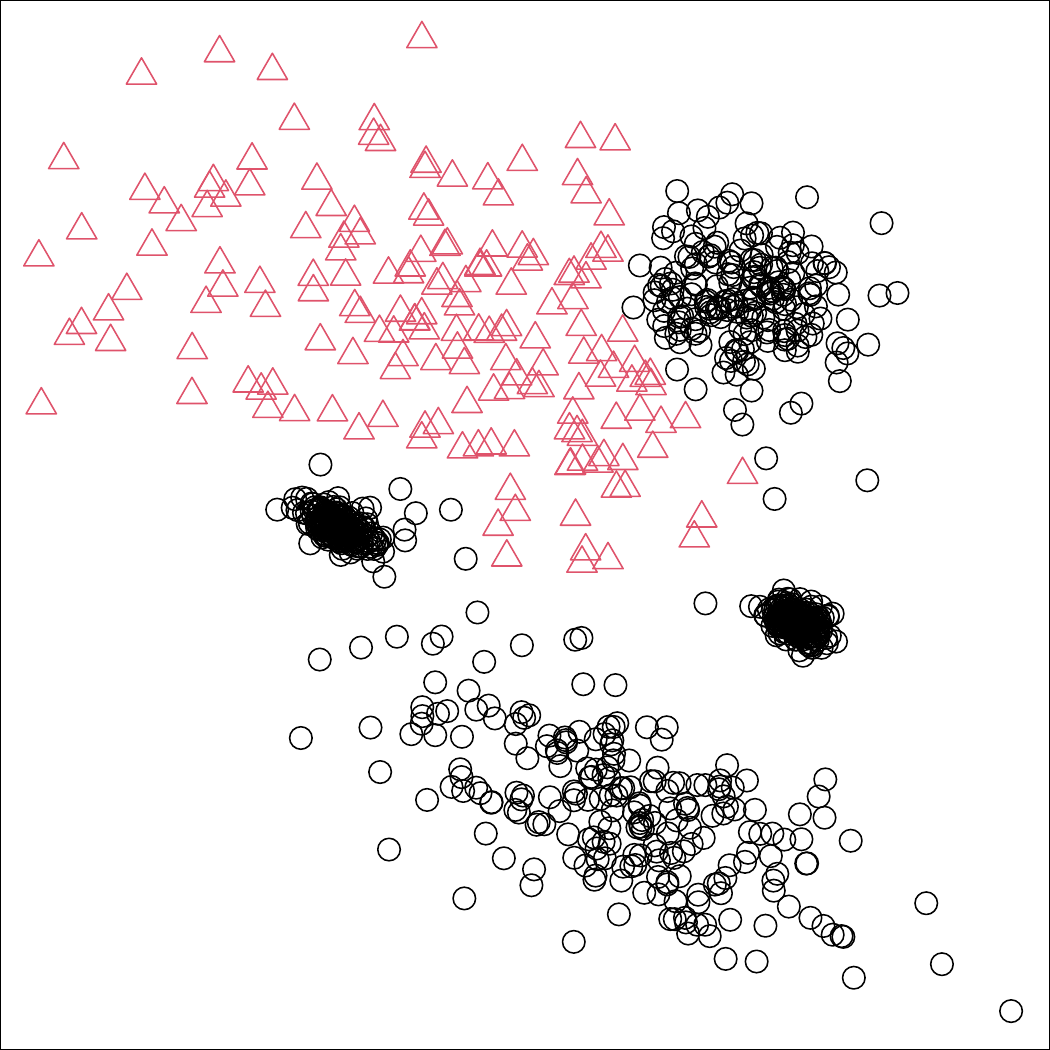}}
    \subfigure[]{\includegraphics[width=0.15\linewidth]{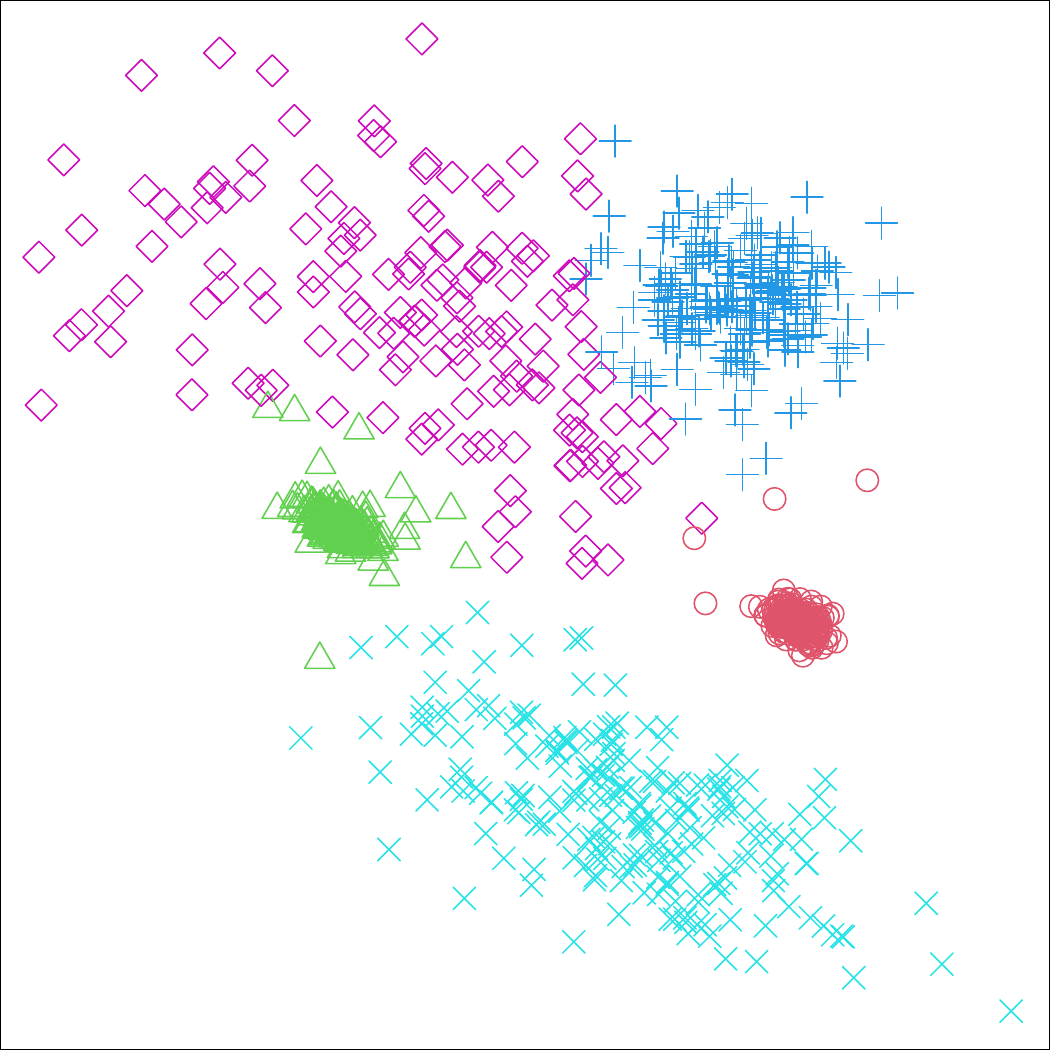}}
    \caption{Equilibrium clusters for $k = 25$ and $\lambda = 2$ (these values were selected automatically using the approach described in Section~\ref{sec:tuning}). (a)-(e) Individual equilibrium clusters; (f) final clustering solution.}
    \label{fig:all_cluster}
\end{figure*}

%
{
\begin{algorithm*}[t]
    \begin{algorithmic}
        \STATE \textit{\#\#\# Initialise cluster counter}
        \STATE $C\gets 0$
        \STATE \textit{\#\#\# Create new equilibrium clusters as long as there is at least one point not yet allocated to any}
        \WHILE{$\min_{j\in [n]} \left| \{j\} \cap \bigcup_{c\in [C]} \C_c\right| = 0$}
        \STATE \textit{\#\#\# Increment cluster counter and find seed of next cluster}
        \STATE $C \gets C+1$
        \STATE $j_C \gets \argmax_{j \in \overline{\bigcup_{c\in [C]}\C_c}} \left|\left\{i \in [n] \big| j \in \nk{\x_i}\right\}\right|$
        \STATE \textit{\#\#\# Grow equilibrium cluster from seed $j_C$}
        \STATE $\C_C \gets \mathrm{EquilibriumCluster}(\sample, j_C, k, \lambda, r, T)$
        \STATE \textit{\#\#\# If seed does not lie in cluster, then create a dummy cluster containing only the seed to avoid re-selection}
        \IF{$j_C \not \in \C_C$}
            \STATE $C \gets C+1$
            \STATE $\C_C \gets \{j_C\}$
        \ENDIF
        \ENDWHILE
        \STATE \textit{\#\#\# Compute membership strengths for all points in all equilibrium clusters}
        \FOR{$i \in [n], c \in [C]$}
            \STATE $s_{i,c} \gets \left(\frac{|\nk{\x_i}\cap \C_c|}{k} - \lambda \frac{|\C_c|}{n}\right)_+$
        \ENDFOR
        \STATE \textit{\#\#\# Allocate each point to the cluster to which it has the greatest membership strength}
        \FOR{$c \in [C]$}
            \STATE $\C^*_c \gets \left\{i \in [n] \big| s_{i,c} = \max_{d \in [C]} s_{i,d}\right\}$
        \ENDFOR
        \RETURN $\{\C^*_c\}_{c\in [C]}$
    \end{algorithmic}
    \caption{A complete clustering algorithm}\label{alg:cluster_all}
\end{algorithm*}}
Figure~\ref{fig:all_cluster} shows the results of applying Algorithm~\ref{alg:cluster_all} to the two-dimensional data set described in the previous subsection. Figures~\ref{fig:all_cluster} (a)-(e) show the five equilibrium clusters obtained, while Figure~\ref{fig:all_cluster} (f) shows the final assignment, which accurately captures all of the clusters in the data. Notice that in this example there is very little overlap in the equilibrium clusters. However, in some cases it is possible that there may be considerably more overlap, but nonetheless the final allocation is able to resolve this appropriately. In addition, as we describe in the following subsection, the most appropriate setting for $\lambda$ is typically that which leads to the least overlap, and we have used this observation in how we select $\lambda$.

\subsection{Setting $\lambda$ and $k$}\label{sec:tuning}

For a given setting of $\lambda$ and $k$, the algorithm described in the previous subsection automatically determines the number of clusters in the final solution. Methods which automatically determine the number of clusters are desirable, since choosing the number of clusters to extract from a given data set is generally seen as one of the most challenging tasks. However, determining settings for any tuning parameters is similarly important, since inappropriate settings may lead to a poor representation of the actual clusters in the data, and so whether or not the number of clusters extracted was determined appropriately for the settings used or not is moot.\\
\\
In this subsection we describe a very simple criterion for selecting values for $\lambda$ and $k$. It is nonetheless worth discussing briefly the effect which varying each of these parameters typically has on the clustering solution. The effect of $\lambda$ may be intuited directly by considering the equilibrium condition in Ineq.~(\ref{eq:equilibrium}). Simply put, as $\lambda$ increases the condition for membership to an equilibrium cluster becomes stricter, generally leading to smaller equilibrium clusters and a greater number of clusters overall required to ``cover'' the data. 
It is worth noting that if $\lambda < 1$ then the entire data set is an equilibrium cluster. However, this does not exclude the possibility that non-trivial subsets of the data can still be equilibrium clusters despite a setting of $\lambda < 1$. Nonetheless, we typically do not explore these values for $\lambda$ and use $\lambda = 1$ as a lower bound in practice.

The value for $k$, the number of neighbours, may be interpreted in terms of the degree of flexibility of the cluster structure which can be identified by the method. Specifically, smaller values for $k$ allow for more detailed \textit{local} structure to be captured, whereas larger values can have the effect of smoothing over the local structure in the data. Smaller values for $k$, however, have the drawback of potentially picking up on spurious patterns which are not inherent to the natural groupings of points, but rather a factor of sampling variation or ``noise''.\\
\\
We have found that using the very simple criterion which maximises the average normalised cluster membership strength to be remarkably effective. That is, if $\{s_{i,c}^{(\lambda, k)}\}_{i \in [n], c \in [C^{(\lambda, k)}]}$ are the equilibrium cluster membership strengths in the solution for a given value for $\lambda$ and $k$, then we select the values
\begin{align*}
    (\lambda^*, k^*) &= \argmax_{(\lambda, k)} \frac{1}{n}\sum_{i=1}^n \frac{\max_c s_{i,c}^{(\lambda, k)}}{\sum_{c=1}^{C^{(\lambda, k)}}s_{i,c}^{(\lambda, k)}}.
\end{align*}
Naturally we wish to select parameters which lead to solutions in which points have strong alignment with their respective cluster assignments. However, because of the possibility of overlap in the equilibrium clusters, it is possible that inappropriate settings may lead to points having high membership strengths to multiple equilibrium clusters. This is why it is important to normalise the raw membership strengths of points to their clusters, i.e. the values $\left\{\max_c s_{i,c}^{(\lambda, k)}\right\}_{i \in [n]}$, to account for this.

\begin{figure*}[t]
    \centering
    \subfigure[Inappropriately small value for $\lambda$ ($\lambda = 1)$.]{\includegraphics[width=0.2\linewidth]{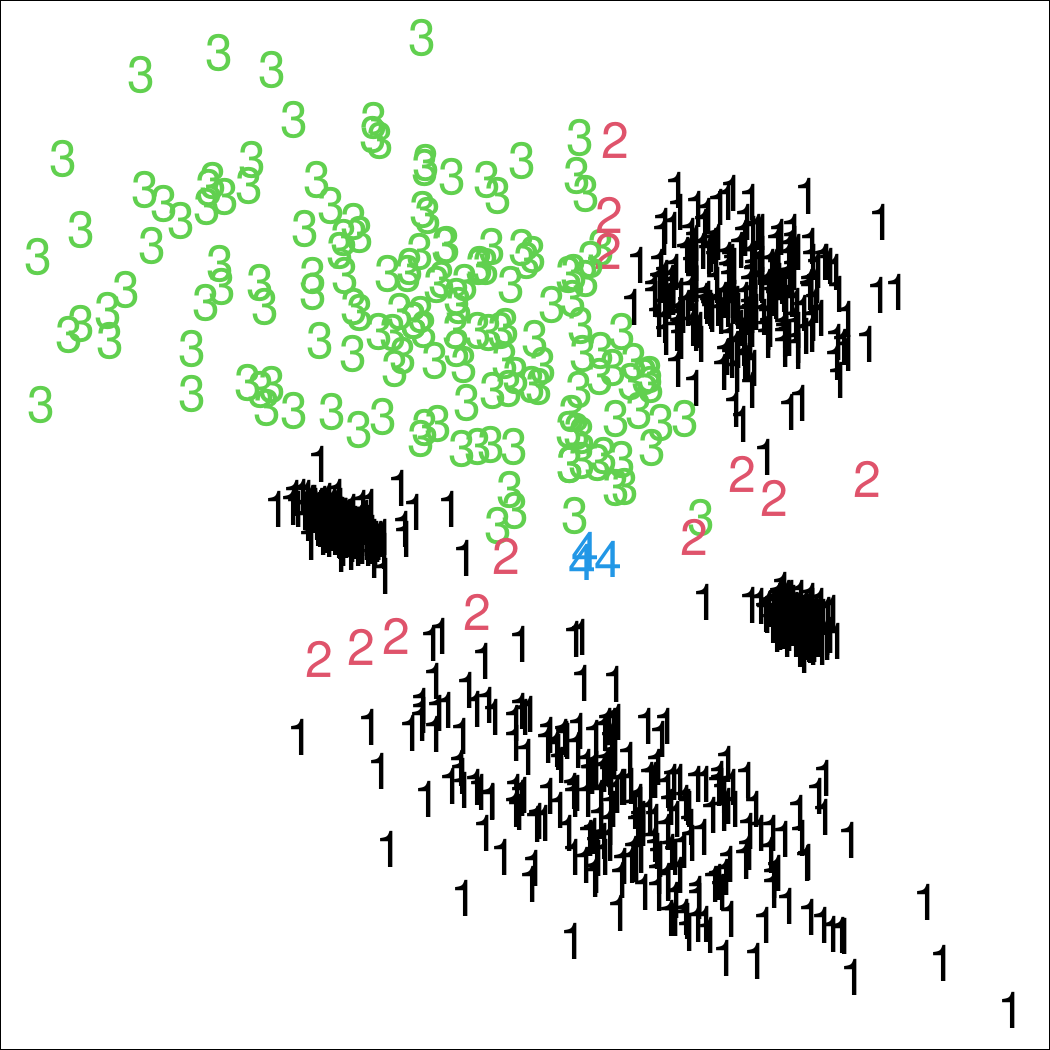} \hspace{1pt}\includegraphics[width=0.2\linewidth]{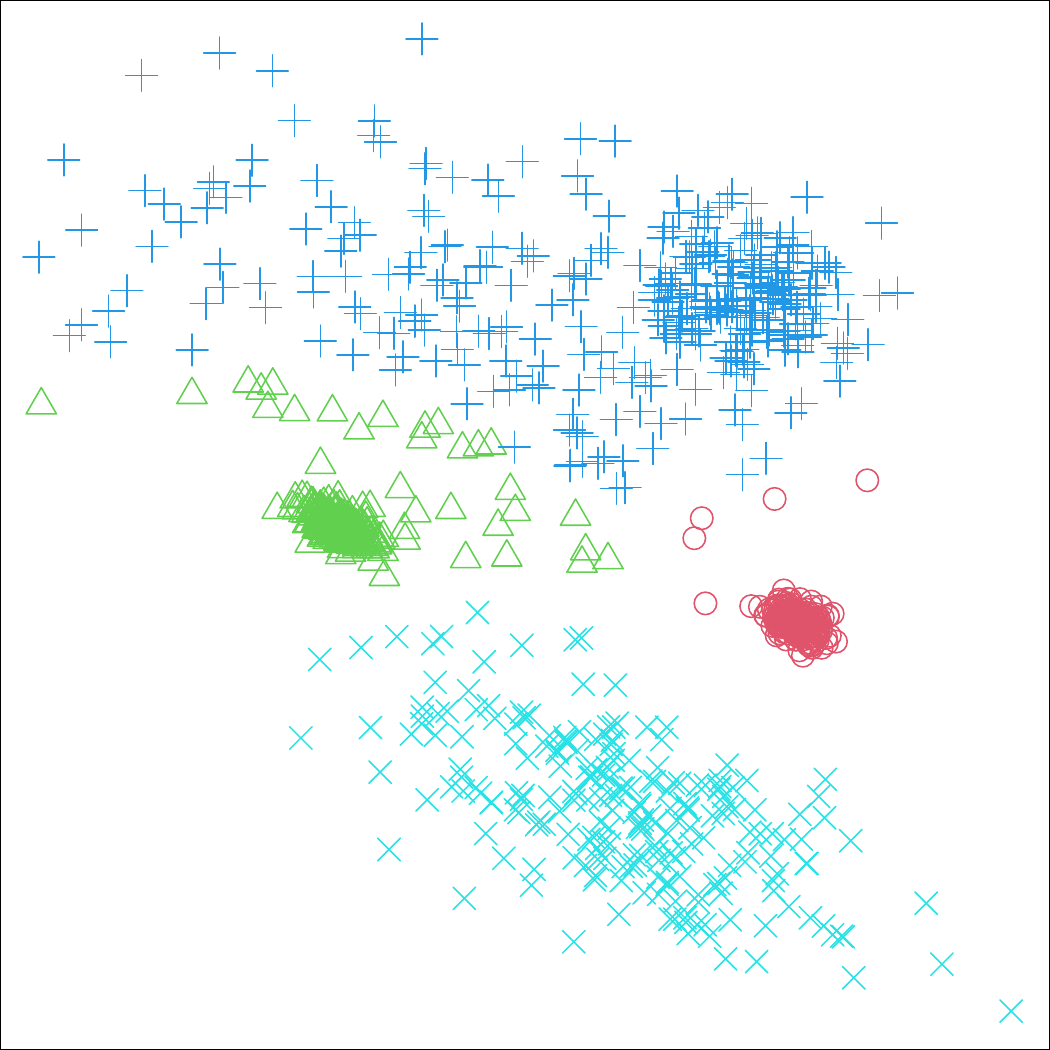}\label{fig:wrongLambda1}}
    \subfigure[Inappropriately large value for $\lambda$ ($\lambda = 3$).]{\includegraphics[width=0.2\linewidth]{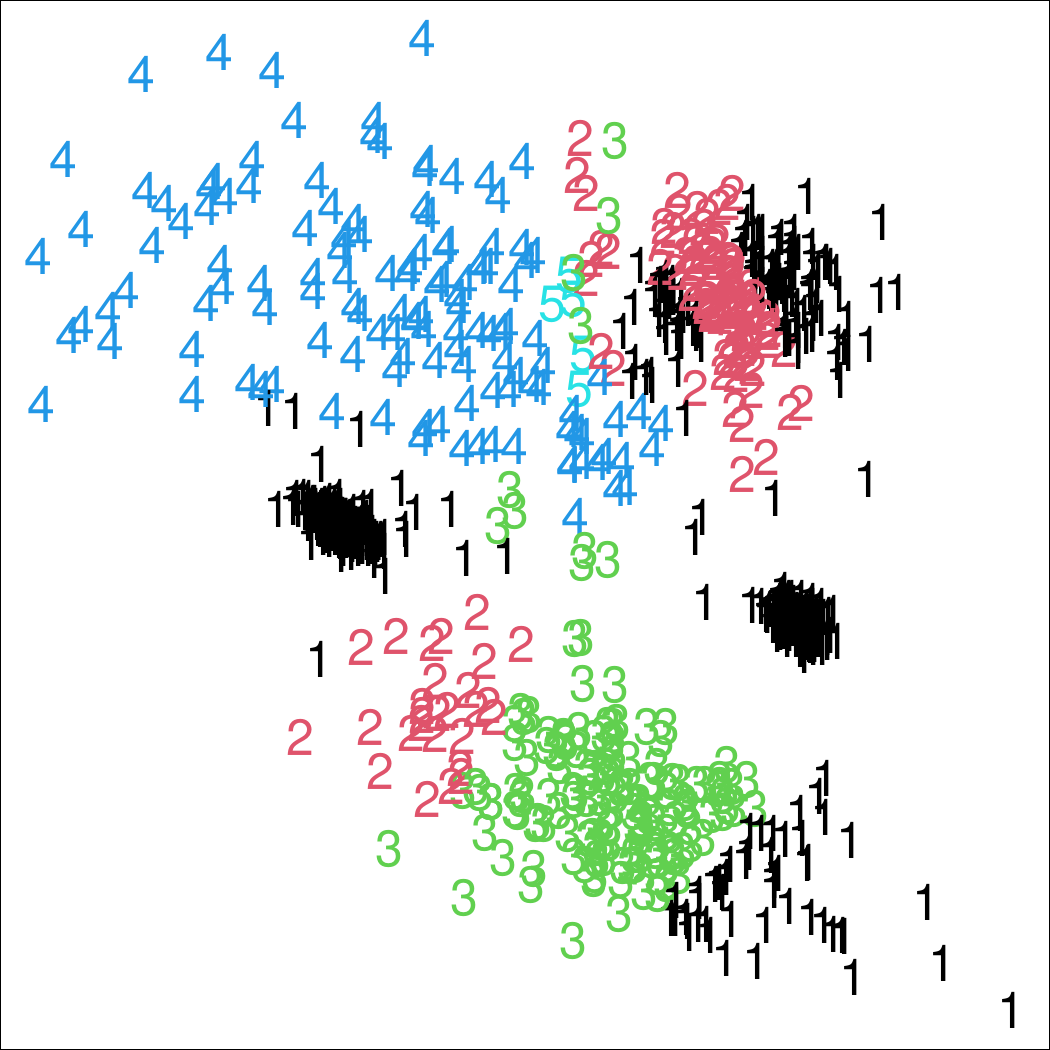} \hspace{1pt}\includegraphics[width=0.2\linewidth]{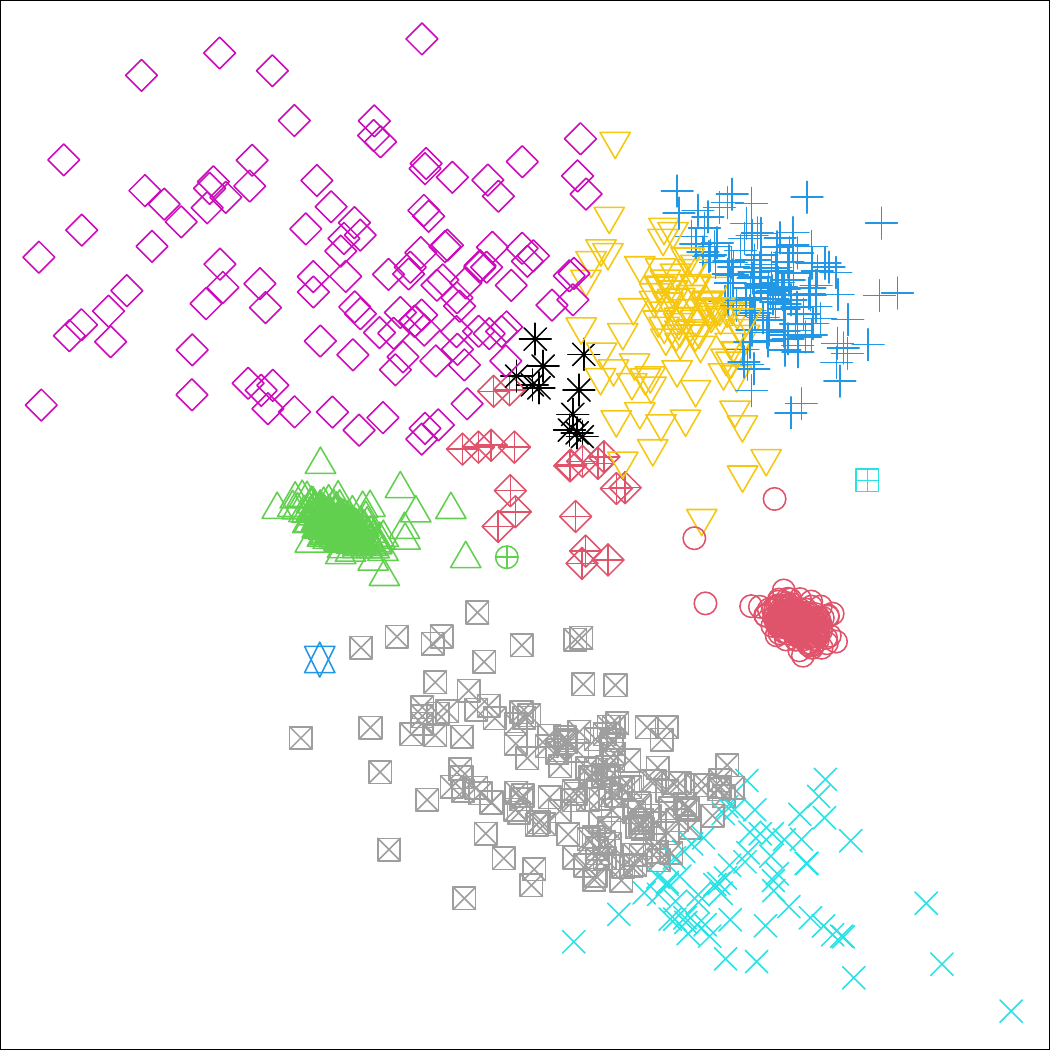}\label{fig:wrongLambda2}}
    \caption{Effect of inappropriate selection of $\lambda$. In each sub-figure the left plot shows the number of equilibruim clusters each point belongs to, while the right plot shows the induced clustering solution.
    Both inapproptiately large \textit{and} small values for $\lambda$ lead to a high degree of overlap in the equilibrium clusters.
    }
    \label{fig:wrongLambda}
\end{figure*}

We refer again to the simple two-dimensional example, from before, for illustration. Recall, in relation to Figure~\ref{fig:all_cluster}, that for the setting of $\lambda = 2$ each of the equilibrium clusters aligned well with what we would see as the ``true'' clusters, and it should be clear that there is very little overlap between them.
On the other hand, both inappropriately large \textit{and} inappropriately small values frequently lead to a larger proportion of overlap. Inappropriately small values for $\lambda$ lead to less strict criteria for equilibrium cluster membership, which directly leads to a higher proportion of the points whose cluster membership may be seen as more ambiguous being included in each equilibrium cluster. An inappropriately large value for $\lambda$ will instead typically lead to an inappropriately large number of equilibrium clusters, and hence a higher proportion of the sample in their shared boundaries. These effects are shown in Figure~\ref{fig:wrongLambda}. In each subfigure the left plot shows the points plotted with numeric characters indicating the numbers of equilibrium clusters to which they each belong, while the right plot shows the final clustering assignment.
The solution with an inappropriately small setting of $\lambda = 1$, Figure~\ref{fig:wrongLambda1}, has four equilibrium clusters, and with the entirety of the most dispersed ``true'' cluster belonging to three equilibrium clusters. In addition, a few points near the centre belong to all four equilibrium clusters. 
The solution with an inappropriately large setting of $\lambda = 3$, Figure~\ref{fig:wrongLambda2}, has fourteen equilibrium clusters. This results in multiple of the ``true'' clusters split, and numerous of their points belonging to multiple equilibrium clusters. We chose these particular settings as in our experiments we only search over $\lambda \in [1, 3]$, and so these represent the extremes of what may have resulted during the search for an appropriate setting.

\section{Experiments}\label{sec:experiments}

In this section we report the results from a large set of experiments with the proposed approach. For context and for the sake of comparison, we also report the results from a number of other clustering methods, both classical and recent.

\subsection{Data Sets}

We investigate the performance of the proposed approach and the existing approaches, on 45 data sets from the public domain which have been used frequently within the clustering literature. The vast majority of these are available from the UCI machine learning repository~\citep{UCI}\footnote{The exceptions are the Yale faces data set (yale), taken from~\url{http://cvc.cs.yale.edu/cvc/projects/yalefacesB/yalefacesB.html} and the yeast data set, originally obtained from~\url{http://genome-www.stanford.edu/cellcycle/}.}. As is common, we use data sets for which ground truth label sets are available, as they allow us to evaluate the performance of clustering methods by their ability to accurately align their clustering solutions with these ground truth clusters. 
Details of all data sets used, in terms of size and numbers of clusters, can be seen in Table~\ref{tb:metadata}. Two of the data sets\footnote{The oliveoil data set has two, and the frogs data set three.} have multiple label sets, and in our comparisons we simply treat all of these as separate clustering problems, leading to a total of 48. It is worth pointing out that this is a far larger collection of data sets than is typically found in such studies.

We used a single pre-processing rule in which we standardised each variable to have unit variance, and when the data contained more than 100 variables in total they were then projected onto their first 100 principal component directions. The reason for this reduction in dimension is purely computational, where, in particular, fitting a large number of Gaussian Mixture Models (one of the methods with which we compare) over very high dimensional data sets can be computationally demanding.

\begin{table*}[]
    \centering
    \footnotesize
    \begin{tabular}{lrrr|lrrr|lrrr}
Data set & $n$ & $d$ & $C$ & Data set & $n$ & $d$ & $C$ & Data set & $n$ & $d$ & $C$\\
\hline
pendigits & 10992 & 16 & 10 & ionosphere & 351 & 33 & 2 & vowel & 990 & 10 & 11\\
optidigits & 5620 & 64 & 10 & banknote & 1372 & 4 & 2 & biodeg & 1055 & 41 & 2\\
mfdigits & 2000 & 216 & 10 & dermatology & 366 & 34 & 6 & ecoli & 336 & 7 & 8\\
wine & 178 & 13 & 3 & forest & 523 & 27 & 4 & led & 500 & 7 & 10\\
oliveoil & 572 & 8 & 3/9 & glass & 214 & 9 & 6 & letter & 20000 & 16 & 26\\
auto & 392 & 7 & 3 & heartdisease & 294 & 13 & 2 & sonar & 208 & 60 & 2\\
yeast & 698 & 72 & 5 & iris & 150 & 4 & 3 & vehicle & 846 & 18 & 4\\
yeast (UCI) & 1484 & 8 & 10 & libra & 360 & 90 & 15 & wdbc & 569 & 30 & 2\\
satellite & 6435 & 36 & 6 & parkinsons & 195 & 22 & 2 & wine & 1599 & 11 & 6\\
seeds & 210 & 7 & 3 & phoneme & 4509 & 256 & 5 & zoo & 101 & 16 & 7\\
imageseg & 2310 & 19 & 7 & votes & 434 & 16 & 2 & dna & 2000 & 180 & 3\\
mammography & 828 & 5 & 2 & frogs & 7195 & 22 & 4/8/10 & msplice & 3175 & 240 & 3\\
breastcancer & 699 & 9 & 2 & isolet & 6238 & 617 & 26 & musk & 6598 & 166 & 2\\
texture & 5500 & 40 & 11 & smartphone & 10929 & 561 & 12 & pima & 768 & 8 & 2\\
soybeans & 683 & 35 & 19 & yale & 5850 & 1200 & 10 & spambase & 4601 & 57 & 2\\

    \end{tabular}
    \caption{List of data sets and their characteristics}
    \label{tb:metadata}
\end{table*}

\subsection{Clustering Methods}

Here we describe all of the clustering methods used in our comparative study. We have included a number of classical approaches, some which are at least a decade old but which have become extremely popular, as well as three more recent approaches. A list of all clustering methods considered, and the approaches we used for model selection, is given below:
\begin{enumerate}
    \item $K$-means (KM): The classical clustering model, in which clusters are allocated to their nearest cluster centroid and centroids are chosen to minimise the sum of squared distances from the points to their nearest centroid. We used the implementation in the {\tt R}~\citep{R} package {\tt ClusterR}~\citep{ClusterR}, and the popular $K$-means++ initialisation~\citep{ArthurV2007}. We used 10 initialisations due to the randomness in the $K$-means++ method, and selected the number of clusters (from 2 to 30) using the silhouette score~\citep{kaufman2009finding}.
    \item Gaussian Mixture Model (GMM): Arguably the most popular model-based clustering model, in which the data distribution is modelled using a Gaussian mixture with each component representing a cluster. We used the implementation in the {\tt ClusterR} package, which uses a $K$-means initialisation before applying a standard EM algorithm to maximise the likelihood. We selected the number of clusters (from 1 to 30) using the Bayesian Information Criterion~\citep{schwarz1978estimating}.
    \item Spectral clustering (SC): We used the approach described by~\cite{NNSC}, which uses a nearest neighbours graph and an unsigned Laplacian. To determine the number of clusters (again from 1 to 30), all eigenvectors whose values transcend the value zero (up to a small error margin) are discarded, after which the remaining eigenvectors are checked for overlaps in the induced clustering, and for a pair of such overlapping eigenvectors the one rendering the lower quality partition is discarded. The remaining eigenvectors determine allocations to clusters by taking the maximiser of the absolute values of the eigenvectors over each index. We experimented with a number of SC variants, and found this approach to be both computationally efficient and provide quite consistently good results. We set the number of nearest neighbours equal to $2\lceil \log(n) \rceil$, after some experimentation with various settings showed this to provide the most consistent results.
    \item Mean-shift (MS): The mean-shift clustering procedure applies iterative local averaging, which has been shown to be equivalent to applying gradient ascent on a non-parametric estimate of the density, thereby aligning clusters with the ``basins of attraction'' of the modes of the data density. We used a nearest neighbours mean-shift algorithm, and set the number of neighbours equal to $\lceil \log(n)\rceil$ as this was a setting which yielded the most consistently good results. We also explored the nearest neighbours mode seeking algorithm described by~\cite{nnms}, which has similarities with mean-shift but iterative updates shift each point to the highest density point from among its neighbours. However, the iterative averaging approach achieved considerably better performance overall.
    \item HDBSCAN (HDB): The classical density based clustering method, DBSCAN~\citep{ester1996density}, aligns clusters with the components of a chosen level set of the data density (as estimated with a uniform kernel density estimator). The more recent Hierarchical DBSCAN~\citep{campello2013density} implicitly fits all DBSCAN models for all bandwidths in the kernel density estimate, and extracts clusters from the induced hierarchical model using a stability criterion. We set the number of neighbours required to classify a point as a ``high density point'' (i.e., in the level set) to each of $\{5, 7, 9, 11, 13, 15\}$ and selected a solution using the Density Based Clustering Validation criterion~\citep[DBCV]{moulavi2014density}. DBSCAN and its variants do not allocate points not in the neighbourhood of a high density point to clusters, instead classifying them as noise. To make the results comparable with other methods, we merged these points with their nearest clusters. This was performed after selection using DBCV.
    \item Border Peeling Clustering (BPC): The approach of~\cite{BPC} which iteratively trims the points believed to be on the borders of clusters, to reveal the cluster cores. These cores are then clustered based on a graph connectivity condition, and the trimmed points are allocated to one of the cores based on a path produced during the trimming phase. As with DBSCAN this approach allocates some points as outliers, and for a fair comparison we similarly merged these with their nearest clusters. We used the implementation provided by the authors at~\url{https://github.com/nadavbar/BorderPeelingClustering}.
    \item Selective Nearest Neighbours Clustering (SNNC): The approach described by~\cite{SNNC}, which has a very similar structure to BPC, but uses fewer peeling/trimming iterations and a different connectivity condition to recover cluster cores as well as a different criterion to allocate border points. We used the code provided by the authors at~\url{https://github.com/SSouhardya/SNNC}. This method could not be run on the Letters data set, due to size, and so we ran it on 10 random samples of size 10 000 from the data and report the average performance.
    \item Torque Clustering (TC): The approach introduced by~\cite{torqueCluster}, which generates a hierarchical clustering model by iteratively merging clusters with their nearest cluster of greater size (or ``mass''). It then extracts the final solution based on the properties of these cluster mergers, in terms of the products of the masses of the clusters being merged and the squared distances between them. We used the implementation provided by the authors at~\url{https://github.com/JieYangBruce/TorqueClustering}, and report the performance using both Euclidean (TCe) and cosine distances (TCc). 
    \item Nearest Neighbour Equilibrium Clustering (NNEC): The proposed approach. We fit models using Algorithm~\ref{alg:cluster_all} for $k \in \{10, 15, 20, 25\}$ and $\lambda \in \{1.0, 1.2, 1.4, ..., 2.8, 3.0\}$, and select a model using the criterion described in the previous section. In our experience any reasonable setting of $k$ can be used, and that more benefit is gained by carefully tuning the value of $\lambda$. However, we have observed that inappropriately small settings of $k$ sometimes cannot be resolved by a conditionally appropriate setting of $\lambda$. When dealing with a data set on which all 44 elements of the grid $\{10, 15, 20, 25\} \times \{1.0, 1.2, ..., 3.0\}$ represents a significant computational burden, we recommend fixing $k = 2\lceil \log(n) \rceil$, as we have seldom seen this setting to be inappropriately small, and then searching over a grid of values in $[0,1]$ of chosen granularity for $\lambda$. For the interested reader, we also experimented with this setting for $k$ and initially selecting $\lambda^*$ from a grid of $[1, 3]$ of length 7 (i.e. in increments of 1/3), and then conducting a refined search on the interval $[\lambda^* - 1/3, \lambda^* + 1/3]$ using a grid of size 10. This reduced the total number of solutions from 44 to 17. Although the performance was, on average, slightly lower than when tuning over both $k$ and $\lambda$, these settings still leaded to better performance than any of the competing methods using all metrics for comparison which we consider.
\end{enumerate}

\subsection{Clustering Results}

\begin{figure*}[t]
    \centering
    \subfigure[Adjusted Mutual Information (Rank$^{\mathrm{AMI}}$)]{\includegraphics[width=.55\linewidth]{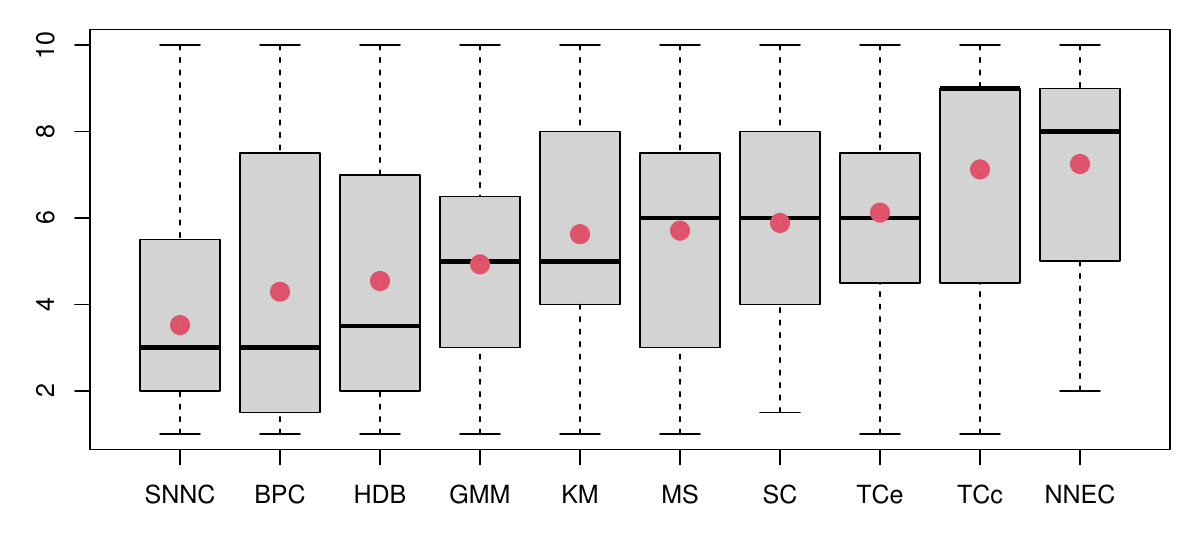}}
    \subfigure[Adjusted Rand Index (Rank$^{\mathrm{ARI}}$)]{\includegraphics[width=.55\linewidth]{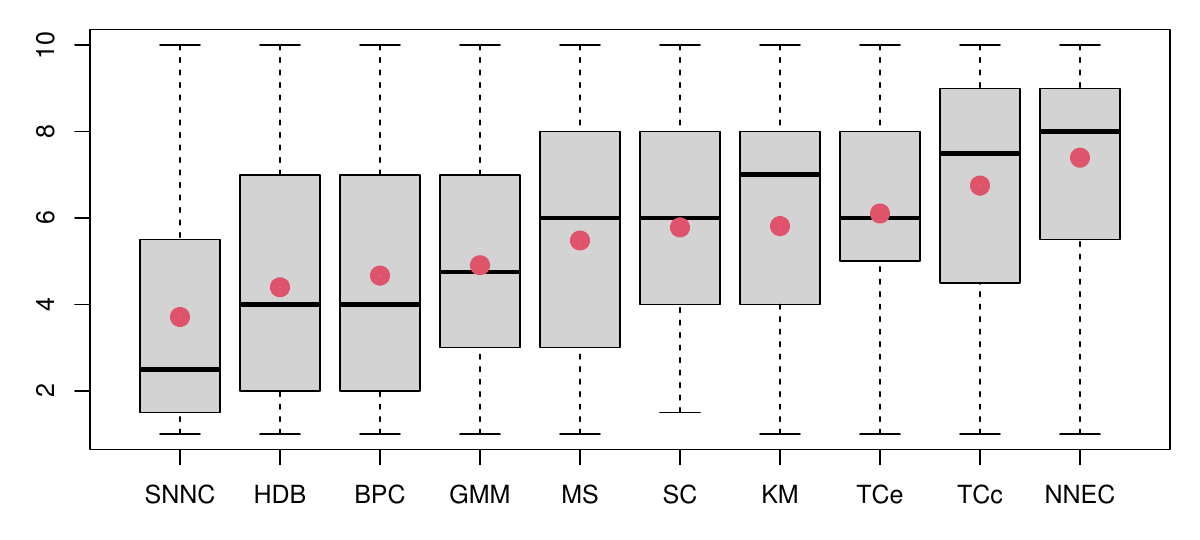}}
    \subfigure[Accuracy (Rank$^{\mathrm{Accuracy}}$)]{\includegraphics[width=.55\linewidth]{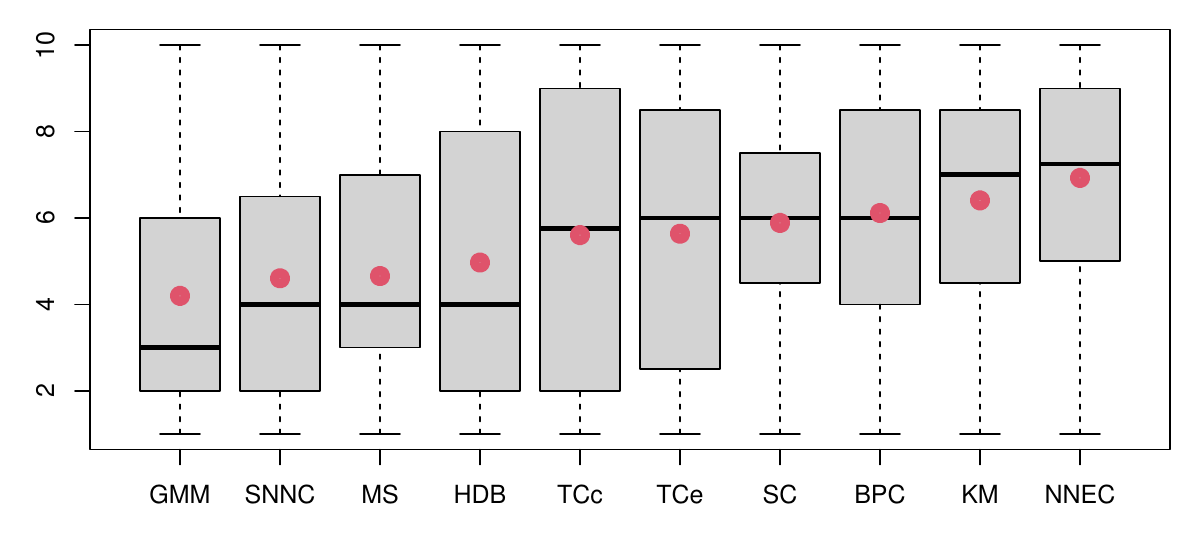}}
    \caption{Distributions of performance ranks across all data sets.}
    \label{fig:rank}
\end{figure*}

\begin{figure*}[t]
    \centering
    \subfigure[Adjusted Mutual Information (AMI$^*$)]{\includegraphics[width=.55\linewidth]{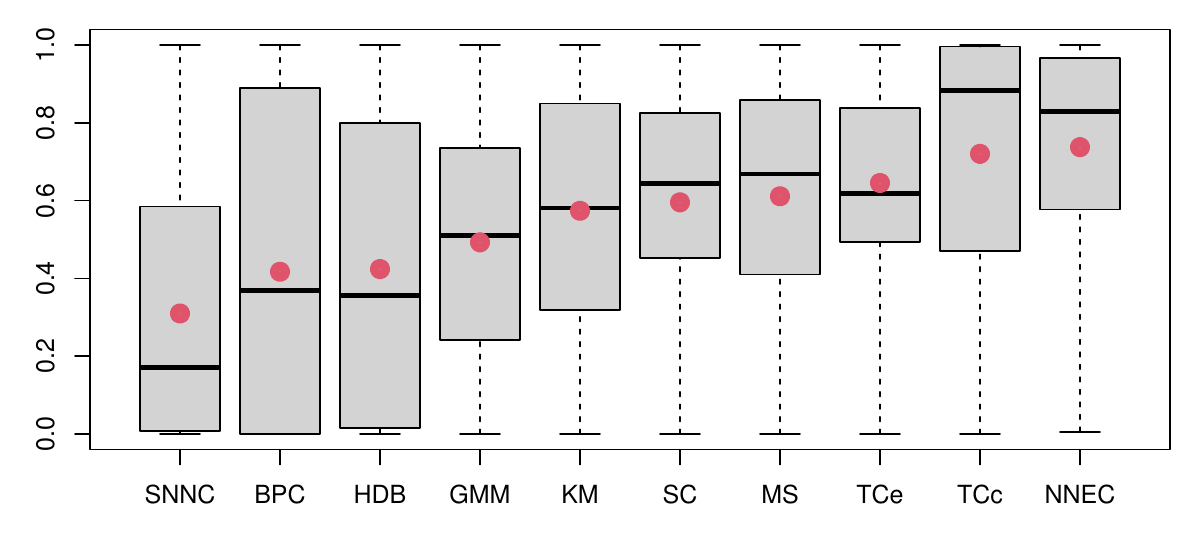}}
    \subfigure[Adjusted Rand Index (ARI$^*$)]{\includegraphics[width=.55\linewidth]{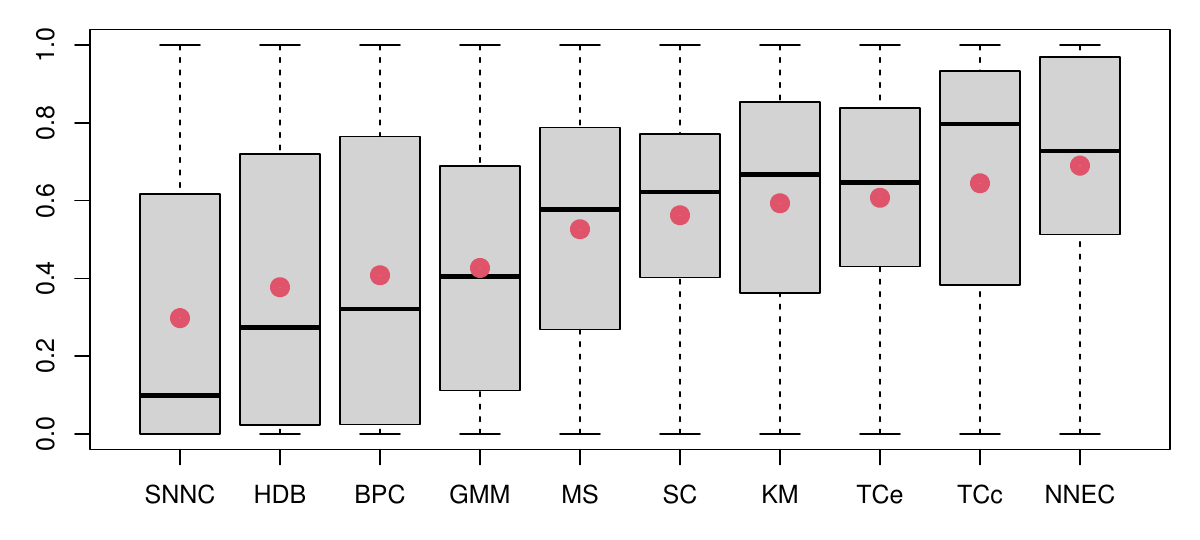}}
    \subfigure[Accuracy (Accuracy$^*$)]{\includegraphics[width=.55\linewidth]{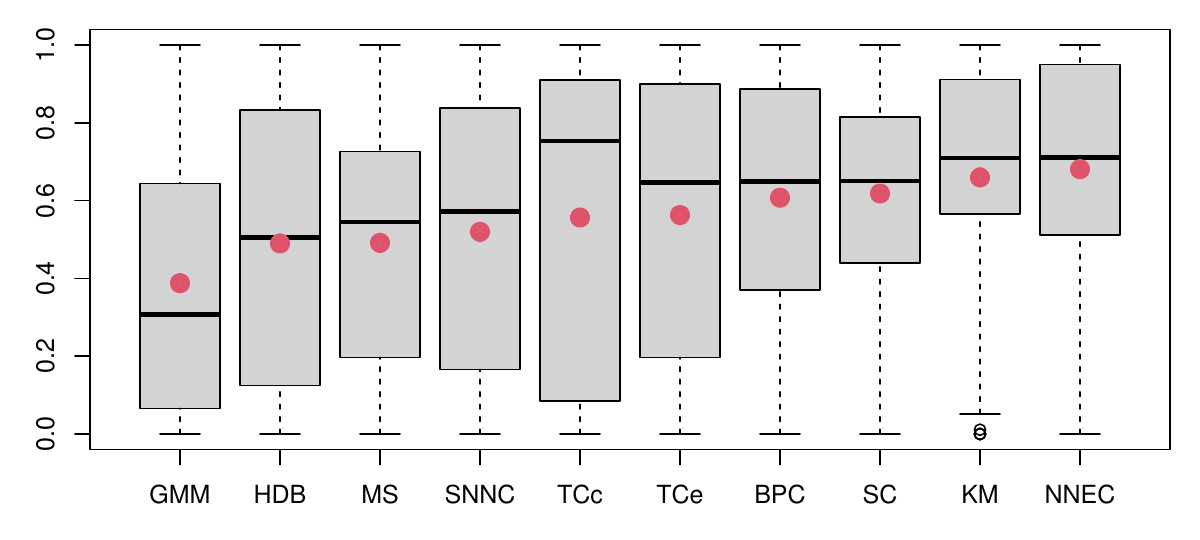}}
    \caption{Distributions of $[0,1]$ mapped performance across all data sets}
    \label{fig:map01}
\end{figure*}

\begin{figure*}[t]
    \centering
    \subfigure[Adjusted Mutual Information (AMI$^{**}$)]{\includegraphics[width=.55\linewidth]{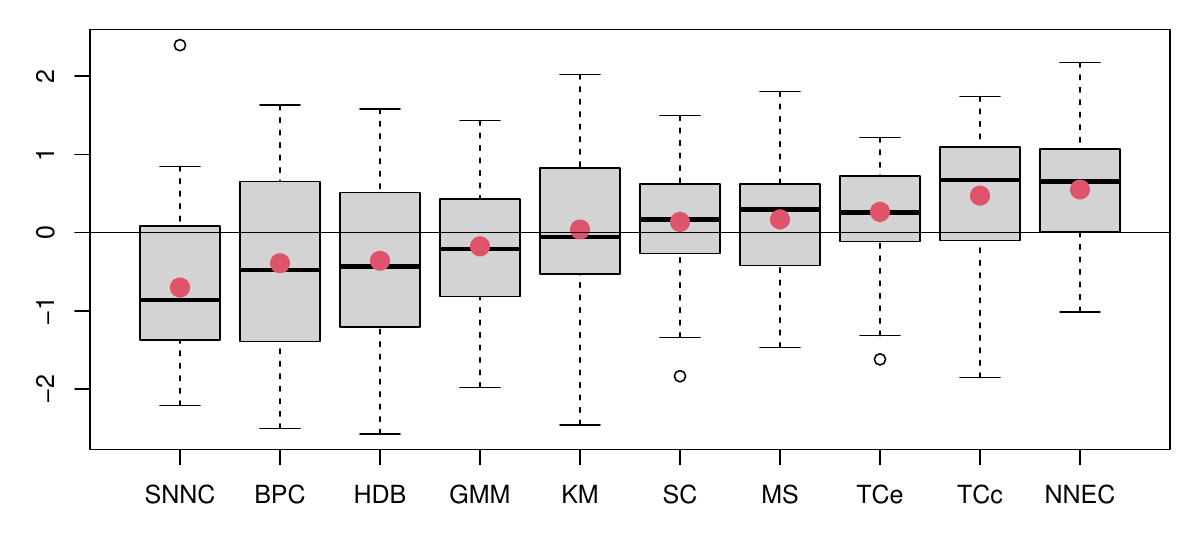}}
    \subfigure[Adjusted Rand Index (ARI$^{**}$)]{\includegraphics[width=.55\linewidth]{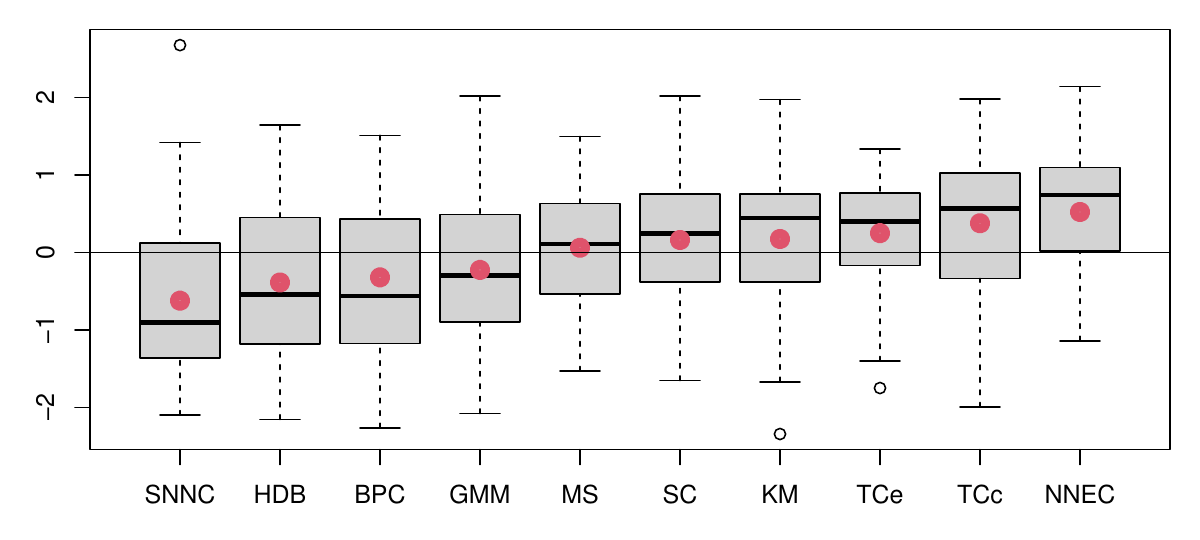}}
    \subfigure[Accuracy (Accuracy$^{**}$)]{\includegraphics[width=.55\linewidth]{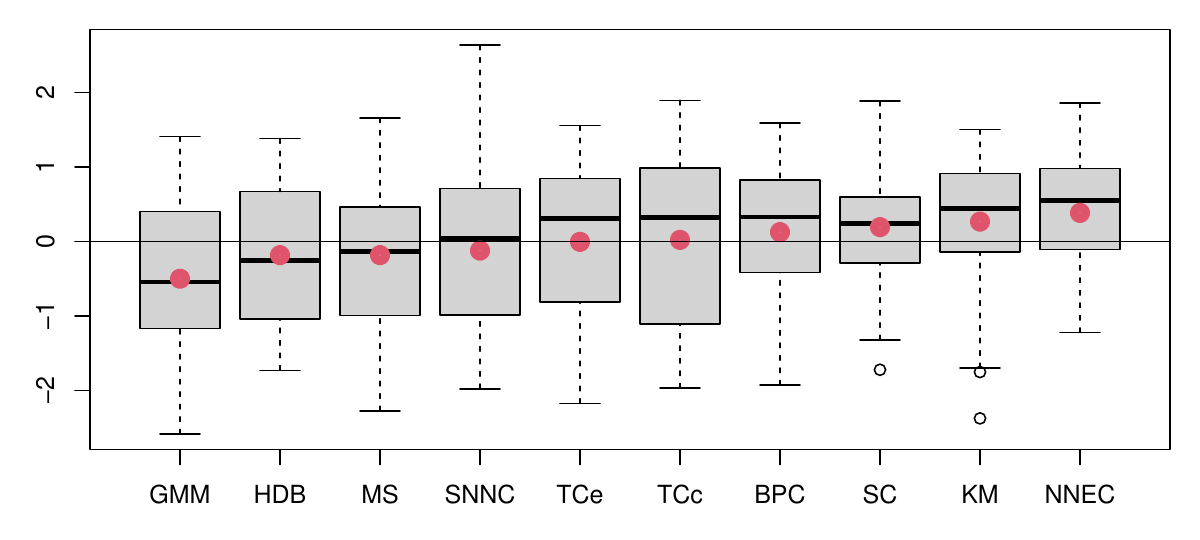}}
    \caption{Distributions of studentised performance across all data sets}
    \label{fig:studentised}
\end{figure*}

\begin{table*}[t]
\centering
\scalebox{.8}{
    
    \begin{tabular}{l|rrrrrrrrrrrrrrrrrrrrrrrrrrrrrrrrrrrrrrrrrrrrrrrr}
          & 
\rotatebox{90}{pendigits} & 
\rotatebox{90}{optidigits} & 
\rotatebox{90}{mfdigits} & 
\rotatebox{90}{wine} & 
\rotatebox{90}{oliveoil ($C=3$)} & 
\rotatebox{90}{oliveoil ($C=9$)} & 
\rotatebox{90}{auto} & 
\rotatebox{90}{yeast} & 
\rotatebox{90}{yeast (UCI)} & 
\rotatebox{90}{satellite} & 
\rotatebox{90}{seeds} & 
\rotatebox{90}{imageseg} & 
\rotatebox{90}{mammography} & 
\rotatebox{90}{breastcancer} & 
\rotatebox{90}{texture} & 
\rotatebox{90}{soybeans}\\
\hline
GMM & 60.65 & 49.06 & 34.46 & 30.32 & 36.88 & 10.87 & 28.56 & 9.79 & 54.32 & 25.89 & 39.17 & 10.75 & 20.49 & {\bf 84.74} & 55.26 & 21.56\\
KM & 62.03 & 66.07 & 63.33 & {\bf 87.16} & 58.62 & 16.53 & 34.62 & 18.17 & 32.11 & 43.99 & 53.40 & {\bf 21.75} & 73.98 & 18.56 & 13.17 & 20.67\\
SC & 70.71 & 67.57 & 76.65 & 37.88 & 51.96 & 24.11 & 29.08 & 20.26 & 59.27 & 45.84 & 52.15 & 10.22 & 26.47 & 75.30 & 61.36 & 5.43\\
HDB & 77.29 & 0.26 & 0.30 & 48.46 & 66.61 & 24.46 & 38.24 & 2.58 & 15.04 & 46.50 & 54.97 & 10.33 & 20.26 & 35.99 & {\bf 69.78} & 6.30\\
MS & 61.75 & 73.39 & 71.22 & 79.28 & 50.03 & 16.19 & 52.84 & 17.18 & 43.43 & 51.24 & 40.97 & 8.16 & 19.04 & 62.93 & 61.68 & 18.58\\
SNNC & 73.80 & 15.29 & 23.34 & -0.38 & 68.51 & 21.45 & 0.49 & 2.30 & 15.57 & -0.00 & 47.45 & 14.33 & 59.52 & 38.33 & 52.48 & 14.57\\
BPC & 73.84 & 0.70 & 81.53 & 0.00 & {\bf 76.14} & 24.34 & 0.00 & 0.00 & 51.95 & 42.03 & 50.20 & 20.37 & 76.96 & 64.70 & 50.86 & 0.00\\
TCe & 58.19 & 68.46 & 82.56 & 74.80 & 61.48 & 18.83 & 25.00 & 11.33 & 57.97 & 40.57 & {\bf 60.01} & 3.76 & 67.12 & 73.79 & 65.69 & 11.94\\
TCc & 81.57 & 78.80 & 84.01 & 72.30 & 71.71 & {\bf 24.98} & 19.53 & 9.29 & 20.68 & 61.28 & 59.20 & 21.63 & {\bf 78.45} & 32.61 & 68.97 & 17.91\\
NNEC & {\bf 84.00} & {\bf 84.02} & {\bf 86.51} & 80.20 & 47.89 & 19.01 & {\bf 56.19} & {\bf 22.62} & {\bf 69.65} & {\bf 72.57} & 56.36 & 15.69 & 31.22 & 75.12 & 43.44 & {\bf 24.08}\\
\hline
&\rotatebox{90}{ionosphere} & 
\rotatebox{90}{banknote} & 
\rotatebox{90}{dermatology} & 
\rotatebox{90}{forest} & 
\rotatebox{90}{glass} & 
\rotatebox{90}{heartdisease} & 
\rotatebox{90}{iris} & 
\rotatebox{90}{libra} & 
\rotatebox{90}{parkinsons} & 
\rotatebox{90}{phoneme} & 
\rotatebox{90}{votes} & 
\rotatebox{90}{frogs ($C=4$)} & 
\rotatebox{90}{frogs ($C=8$)} & 
\rotatebox{90}{frogs ($C=10$)} & 
\rotatebox{90}{isolet} & 
\rotatebox{90}{smartphone}\\
\hline
GMM & 20.05 & 82.76 & 19.21 & {\bf 33.14} & 11.78 & 27.57 & 42.84 & 9.33 & 14.19 & 16.95 & 22.90 & 38.11 & 42.78 & 79.87 & 63.44 & 33.10\\
KM & 19.32 & 68.90 & 21.15 & 19.22 & 14.50 & 57.68 & 52.59 & 8.14 & 47.35 & {\bf 47.92} & 38.01 & 16.27 & 33.73 & 67.30 & 67.25 & 34.20\\
SC & 22.66 & 83.33 & 34.29 & 31.91 & 14.28 & 28.06 & 38.52 & 14.68 & 55.17 & 16.51 & 28.97 & 40.43 & 38.68 & 78.29 & 73.47 & 41.74\\
HDB & 16.21 & 65.61 & 15.30 & 15.39 & -0.17 & 57.68 & 9.49 & 20.90 & 73.86 & 14.44 & {\bf 38.42} & 1.27 & 4.11 & 14.03 & 66.45 & {\bf 55.17}\\
MS & 17.54 & 80.54 & 32.04 & 25.55 & 13.55 & 52.05 & 46.73 & 12.75 & 80.08 & 14.09 & 23.51 & {\bf 71.75} & 48.93 & 51.50 & {\bf 82.69} & 33.36\\
SNNC & {\bf 39.89} & 64.68 & 2.10 & 25.75 & -0.21 & 57.39 & 20.74 & -1.03 & 0.86 & 27.63 & 17.64 & 5.70 & 47.20 & 40.34 & 63.31 & 15.97\\
BPC & 14.07 & 40.73 & 0.00 & 16.53 & 0.00 & 57.68 & 0.00 & 0.00 & {\bf 80.26} & 45.76 & 36.61 & 26.69 & 42.04 & 77.03 & 59.50 & 51.98\\
TCe & 21.17 & 78.64 & 24.97 & 21.38 & 10.74 & 57.68 & {\bf 58.8} & 9.39 & 79.12 & 23.51 & 34.43 & 56.50 & 41.64 & {\bf 86.32} & 72.66 & 51.78\\
TCc & 26.57 & {\bf 91.31} & {\bf 37.49} & 33.13 & 13.03 & {\bf 63.34} & 55.67 & {\bf 22.61} & 47.16 & 46.47 & 23.65 & 66.57 & 42.80 & 86.20 & 62.73 & 33.93\\
NNEC & 25.85 & 82.55 & 36.36 & 18.32 & {\bf 17.75} & 57.68 & 34.34 & 17.50 & 60.13 & 25.33 & 26.80 & 63.07 & {\bf 55.45} & 83.61 & 78.73 & 38.58\\
\hline
&\rotatebox{90}{yale} & 
\rotatebox{90}{vowel} & 
\rotatebox{90}{biodeg} & 
\rotatebox{90}{ecoli} & 
\rotatebox{90}{led} & 
\rotatebox{90}{letter} & 
\rotatebox{90}{sonar} & 
\rotatebox{90}{vehicle} & 
\rotatebox{90}{wdbc} & 
\rotatebox{90}{wine} & 
\rotatebox{90}{zoo} & 
\rotatebox{90}{dna} & 
\rotatebox{90}{msplice} & 
\rotatebox{90}{musk} & 
\rotatebox{90}{pima} & 
\rotatebox{90}{spambase}\\
\hline
GMM & 45.12 & {\bf 39.92} & 9.06 & 42.84 & 42.57 & 39.73 & 0.00 & 21.32 & 39.59 & 5.71 & 73.83 & {\bf 14.98} & 0.00 & 3.42 & 2.26 & 11.81\\
KM & 46.66 & 37.88 & {\bf 15.24} & 58.06 & 44.81 & 0.69 & 0.30 & 8.98 & 54.37 & 2.75 & 36.65 & 14.81 & 10.73 & 2.89 & {\bf 8.06} & 0.54\\
SC & 58.26 & 20.29 & 7.15 & 21.59 & 44.37 & 37.25 & 6.93 & 17.43 & 0.00 & {\bf 8.61} & 70.84 & 2.27 & 0.00 & 5.40 & 5.06 & {\bf 13.04}\\
HDB & {\bf 74.54} & 36.50 & 1.16 & 5.56 & 42.52 & 47.99 & 2.35 & 0.44 & 41.46 & -0.00 & 67.95 & 0.52 & 6.08 & {\bf 7.52} & 0.51 & 8.03\\
MS & 44.36 & 39.33 & 6.77 & 46.45 & 44.41 & 48.48 & {\bf 9.27} & 15.58 & 42.33 & 7.62 & 65.98 & 4.21 & 4.64 & 5.06 & 2.01 & 8.88\\
SNNC & 12.93 & 0.00 & 2.44 & 38.65 & 53.33 & 0.00 & 4.74 & 3.52 & 0.38 & 0.31 & 72.62 & -0.17 & 0.26 & 1.66 & 1.14 & 5.14\\
BPC & 70.65 & 3.96 & 0.00 & 31.35 & {\bf 60.41} & 49.88 & 0.00 & 10.65 & 0.00 & 0.00 & 0.00 & 0.00 & 0.00 & 6.27 & -0.00 & 2.81\\
TCe & 61.03 & 37.48 & 7.14 & 36.91 & 38.92 & {\bf 54.85} & 6.81 & 16.15 & 38.06 & 4.98 & 76.22 & 7.26 & 4.91 & 4.90 & 3.60 & 6.97\\
TCc & 45.26 & 36.43 & 5.30 & {\bf 63.2} & 28.79 & 51.85 & 5.28 & {\bf 27.37} & {\bf 69.5} & 7.57 & {\bf 76.62} & 6.44 & 7.06 & 0.93 & 4.08 & 8.72\\
NNEC & 51.98 & 24.23 & 8.57 & 55.54 & 56.23 & 45.59 & 0.05 & 12.72 & 60.74 & 8.00 & 70.92 & 12.97 & {\bf 17.53} & 4.79 & 4.51 & 7.82\\
    \end{tabular}}
    \caption{Adjusted Mutual Information for all methods across all data sets}
    \label{tab:nmi_all}
\end{table*}

\begin{table*}[t]
\centering
\scalebox{.8}{
    
    \begin{tabular}{l|rrrrrrrrrrrrrrrrrrrrrrrrrrrrrrrrrrrrrrrrrrrrrrrr}
 & 
\rotatebox{90}{pendigits} & 
\rotatebox{90}{optidigits} & 
\rotatebox{90}{mfdigits} & 
\rotatebox{90}{wine} & 
\rotatebox{90}{oliveoil ($C=3$)} & 
\rotatebox{90}{oliveoil ($C=9$)} & 
\rotatebox{90}{auto} & 
\rotatebox{90}{yeast} & 
\rotatebox{90}{yeast (UCI)} & 
\rotatebox{90}{satellite} & 
\rotatebox{90}{seeds} & 
\rotatebox{90}{imageseg} & 
\rotatebox{90}{mammography} & 
\rotatebox{90}{breastcancer} & 
\rotatebox{90}{texture} & 
\rotatebox{90}{soybeans}\\
\hline
GMM & 51.97 & 32.24 & 24.75 & 18.90 & 25.61 & 2.58 & 17.85 & 0.47 & 51.21 & 13.29 & 30.17 & 14.25 & 22.04 & {\bf 77.17} & 37.04 & 24.24\\
KM & 47.88 & 63.33 & 56.06 & {\bf 89.75} & 58.42 & -4.31 & 41.79 & 13.42 & 29.36 & 48.05 & 46.07 & 31.49 & 84.45 & 11.09 & 4.85 & 27.09\\
SC & 64.74 & 56.61 & 71.55 & 35.77 & 50.07 & 5.67 & 34.21 & 13.17 & 54.58 & 48.81 & 41.46 & 6.81 & 22.86 & 66.73 & 33.00 & 3.14\\
HDB & 76.28 & 0.00 & 0.01 & 47.66 & 78.52 & 5.41 & 44.86 & 1.16 & 8.09 & 48.83 & 50.83 & 10.79 & 19.75 & 14.93 & 37.35 & -4.64\\
MS & 50.46 & 70.29 & 67.31 & 80.01 & 51.67 & 12.37 & 53.55 & {\bf 14.99} & 37.82 & 53.33 & 19.99 & 3.24 & 18.10 & 46.75 & 36.47 & 18.23\\
SNNC & 62.52 & 4.74 & 9.32 & -0.82 & 79.18 & 4.67 & 0.14 & 1.33 & 8.16 & 0.00 & 24.84 & 28.68 & 82.54 & 14.45 & 35.91 & {\bf 33.38}\\
BPC & 71.09 & 0.02 & 77.15 & 0.00 & {\bf 85.51} & 5.85 & 0.00 & 0.00 & 35.10 & 45.95 & 27.20 & 31.78 & 86.61 & 47.97 & 23.09 & 0.00\\
TCe & 36.94 & 57.22 & 76.71 & 74.14 & 62.53 & 14.63 & 29.50 & 11.05 & 59.75 & 45.26 & 51.38 & 0.28 & 78.59 & 61.87 & 41.92 & 19.69\\
TCc & 74.67 & 74.92 & 80.67 & 74.32 & 69.57 & {\bf 24.64} & 6.18 & 10.50 & 20.48 & 62.19 & 51.15 & {\bf 34.83} & {\bf 87.69} & 20.16 & {\bf 48.55} & 23.99\\
NNEC & {\bf 78.07} & {\bf 79.58} & {\bf 85.5} & 81.70 & 38.12 & 14.72 & {\bf 56.47} & 13.75 & {\bf 68.49} & {\bf 77.5} & {\bf 53.87} & 21.78 & 32.44 & 64.45 & 29.48 & 26.65\\
\hline
&\rotatebox{90}{ionosphere} & 
\rotatebox{90}{banknote} & 
\rotatebox{90}{dermatology} & 
\rotatebox{90}{forest} & 
\rotatebox{90}{glass} & 
\rotatebox{90}{heartdisease} & 
\rotatebox{90}{iris} & 
\rotatebox{90}{libra} & 
\rotatebox{90}{parkinsons} & 
\rotatebox{90}{phoneme} & 
\rotatebox{90}{votes} & 
\rotatebox{90}{frogs ($C=4$)} & 
\rotatebox{90}{frogs ($C=8$)} & 
\rotatebox{90}{frogs ($C=10$)} & 
\rotatebox{90}{isolet} & 
\rotatebox{90}{smartphone}\\
\hline
GMM & 7.70 & 79.92 & 16.26 & 20.98 & 16.64 & 16.31 & 28.58 & 3.56 & 14.44 & 9.14 & 9.50 & 18.46 & 30.38 & 69.84 & 47.95 & 12.89\\
KM & 7.72 & 65.39 & 18.24 & 19.30 & {\bf 28.3} & 56.81 & 31.22 & -9.78 & 39.97 & 57.01 & 40.22 & 6.04 & 30.23 & 51.34 & 76.62 & 44.63\\
SC & 12.38 & 81.66 & {\bf 37} & 23.11 & 16.77 & 22.94 & 24.97 & {\bf 23.77} & 52.93 & 14.42 & 34.47 & 19.38 & 27.24 & 72.19 & 78.72 & 44.43\\
HDB & 4.48 & 59.59 & 11.26 & 12.15 & -0.70 & 56.81 & 1.95 & 11.09 & 68.92 & 6.28 & {\bf 54.55} & 0.07 & 1.25 & 3.45 & 62.45 & {\bf 68.38}\\
MS & 5.39 & 78.40 & 17.14 & 11.78 & 17.47 & 49.32 & 28.05 & 8.44 & 70.41 & 8.66 & 10.57 & {\bf 48.21} & 42.76 & 26.69 & {\bf 85.44} & 14.30\\
SNNC & {\bf 52.49} & 55.86 & 1.91 & {\bf 25.77} & -1.61 & 55.84 & 8.21 & -2.63 & -0.21 & 46.18 & 15.46 & 0.41 & 31.76 & 8.63 & 60.16 & 14.38\\
BPC & 7.49 & 28.32 & 0.00 & 14.65 & 0.00 & 56.81 & 0.00 & 0.00 & {\bf 78.42} & {\bf 58.28} & 53.22 & 8.39 & 33.77 & 56.29 & 56.93 & 66.37\\
TCe & 9.84 & 73.16 & 16.07 & 12.40 & 11.68 & 56.81 & {\bf 35.81} & 1.92 & 77.30 & 24.25 & 47.28 & 32.31 & 22.41 & 77.79 & 77.42 & 56.79\\
TCc & 19.13 & {\bf 91.73} & 29.32 & 24.52 & 14.95 & {\bf 58.19} & 35.11 & 7.96 & 43.33 & 52.91 & 8.65 & 41.90 & 24.22 & 73.13 & 70.64 & 11.68\\
NNEC & 17.38 & 82.04 & 26.45 & 11.36 & 19.96 & 56.81 & 22.06 & 12.05 & 57.54 & 28.51 & 15.35 & 43.43 & {\bf 51.17} & {\bf 78.83} & 62.94 & 20.55\\
\hline
&\rotatebox{90}{yale} & 
\rotatebox{90}{vowel} & 
\rotatebox{90}{biodeg} & 
\rotatebox{90}{ecoli} & 
\rotatebox{90}{led} & 
\rotatebox{90}{letter} & 
\rotatebox{90}{sonar} & 
\rotatebox{90}{vehicle} & 
\rotatebox{90}{wdbc} & 
\rotatebox{90}{wine} & 
\rotatebox{90}{zoo} & 
\rotatebox{90}{dna} & 
\rotatebox{90}{msplice} & 
\rotatebox{90}{musk} & 
\rotatebox{90}{pima} & 
\rotatebox{90}{spambase}\\
\hline
GMM & 20.76 & {\bf 19.37} & {\bf 8.7} & 41.22 & 35.98 & 12.94 & 0.00 & 15.22 & 55.75 & 3.09 & {\bf 83.87} & 7.67 & 0.00 & -0.77 & 1.07 & {\bf 10.86}\\
KM & 65.48 & 17.51 & 5.32 & 69.71 & 39.48 & 0.33 & -0.24 & 8.32 & 67.07 & 4.33 & 25.13 & {\bf 19.25} & 14.62 & 3.69 & {\bf 16.01} & -0.49\\
SC & 64.08 & 11.17 & 4.22 & 15.56 & 34.52 & 12.07 & {\bf 9.64} & 12.13 & 0.00 & 2.67 & 81.20 & 0.33 & 0.00 & 3.87 & 10.88 & 7.37\\
HDB & {\bf 91.9} & 13.52 & -2.35 & 3.80 & 35.16 & 21.45 & -0.01 & 0.06 & 47.05 & -0.22 & 52.81 & 0.42 & 3.90 & {\bf 6.72} & 1.46 & 2.26\\
MS & 22.60 & 18.59 & 3.80 & 36.80 & 37.52 & 16.44 & 5.46 & 7.68 & 48.08 & 5.31 & 50.93 & 4.09 & 3.98 & 0.25 & -0.40 & 7.01\\
SNNC & 9.62 & 0.00 & -5.42 & 43.22 & 48.40 & 0.00 & 0.31 & 0.36 & 0.91 & -0.54 & 69.09 & -0.40 & -0.34 & -5.87 & 3.29 & 1.54\\
BPC & 89.75 & 1.21 & 0.00 & 38.51 & {\bf 54.97} & 16.58 & 0.00 & 9.05 & 0.00 & 0.00 & 0.00 & 0.00 & 0.00 & 2.35 & 0.00 & 1.54\\
TCe & 78.16 & 17.63 & 4.07 & 42.19 & 33.45 & {\bf 27.49} & 4.63 & 10.31 & 43.02 & 1.38 & 70.01 & 12.01 & 1.15 & 0.43 & 1.89 & 3.82\\
TCc & 18.77 & 16.04 & 1.93 & {\bf 71.75} & 17.46 & 22.82 & 5.56 & {\bf 22.11} & {\bf 80.51} & 4.44 & 72.58 & 0.36 & 0.26 & 0.02 & 2.71 & 2.79\\
NNEC & 31.86 & 13.17 & 4.64 & 69.86 & 51.05 & 21.76 & -0.15 & 10.27 & 73.06 & {\bf 5.58} & 80.93 & 13.19 & {\bf 18.95} & 1.22 & 11.00 & 2.25\\

    \end{tabular}}
    \caption{Adjusted Rand Index for all methods across all data sets}
    \label{tab:ari_all}
\end{table*}

\begin{table*}[t]
\centering
\scalebox{.8}{
    \begin{tabular}{l|rrrrrrrrrrrrrrrrrrrrrrrrrrrrrrrrrrrrrrrrrrrrrrrr}
 & 
\rotatebox{90}{pendigits} & 
\rotatebox{90}{optidigits} & 
\rotatebox{90}{mfdigits} & 
\rotatebox{90}{wine} & 
\rotatebox{90}{oliveoil ($C=3$)} & 
\rotatebox{90}{oliveoil ($C=9$)} & 
\rotatebox{90}{auto} & 
\rotatebox{90}{yeast} & 
\rotatebox{90}{yeast (UCI)} & 
\rotatebox{90}{satellite} & 
\rotatebox{90}{seeds} & 
\rotatebox{90}{imageseg} & 
\rotatebox{90}{mammography} & 
\rotatebox{90}{breastcancer} & 
\rotatebox{90}{texture} & 
\rotatebox{90}{soybeans}\\
\hline
GMM & 55.06 & 48.45 & 29.70 & 28.09 & 38.81 & 23.47 & 24.07 & 30.59 & 62.64 & 20.95 & 43.51 & 35.75 & 46.35 & {\bf 80.38} & 49.05 & 50.14\\
KM & 60.26 & 70.91 & 69.40 & {\bf 96.63} & 69.93 & 46.94 & 59.03 & 41.85 & 51.70 & 65.71 & 54.55 & {\bf 74.03} & 95.99 & 18.18 & 23.43 & 65.53\\
SC & 70.33 & 67.05 & 81.65 & 60.11 & 59.44 & 45.92 & 58.17 & {\bf 47.51} & 67.38 & 66.67 & 47.79 & 20.77 & 47.50 & 70.65 & 53.44 & 47.29\\
HDB & 78.45 & 10.50 & 10.50 & 65.17 & 78.32 & 46.17 & 63.47 & 31.94 & 33.04 & 64.76 & 56.97 & 28.14 & 39.91 & 36.35 & {\bf 63.1} & 53.28\\
MS & 51.08 & 72.01 & 68.35 & 93.26 & 56.29 & 42.86 & 69.63 & 40.70 & 45.30 & 69.05 & 24.16 & 9.90 & 37.91 & 49.16 & 52.42 & 61.82\\
SNNC & 69.60 & 20.73 & 22.95 & 37.64 & 82.87 & 44.13 & 36.25 & 31.54 & 30.94 & 33.81 & 43.03 & 62.20 & 93.85 & 35.98 & 55.34 & 69.80\\
BPC & 72.78 & 10.94 & 83.60 & 39.89 & {\bf 86.19} & 45.15 & 36.53 & 31.20 & 52.17 & 64.76 & 46.41 & 62.92 & 96.57 & 56.51 & 47.29 & 64.10\\
TCe & 34.82 & 58.31 & 85.20 & 91.01 & 74.83 & 45.92 & 59.17 & 37.20 & 75.04 & 63.33 & {\bf 60.78} & 1.81 & 94.42 & 63.25 & 57.39 & {\bf 72.65}\\
TCc & 81.96 & 79.61 & 91.05 & 91.01 & 82.52 & {\bf 63.27} & 14.61 & 33.15 & 42.50 & 84.29 & 57.23 & 67.03 & {\bf 96.85} & 27.27 & 57.69 & 47.58\\
NNEC & {\bf 87.35} & {\bf 86.98} & {\bf 93.15} & 93.82 & 49.83 & 48.21 & {\bf 73.35} & 40.36 & {\bf 80.14} & {\bf 91.9} & 60.39 & 48.31 & 54.65 & 71.82 & 40.85 & 62.96\\
\hline
 & 
\rotatebox{90}{ionosphere} & 
\rotatebox{90}{banknote} & 
\rotatebox{90}{dermatology} & 
\rotatebox{90}{forest} & 
\rotatebox{90}{glass} & 
\rotatebox{90}{heartdisease} & 
\rotatebox{90}{iris} & 
\rotatebox{90}{libra} & 
\rotatebox{90}{parkinsons} & 
\rotatebox{90}{phoneme} & 
\rotatebox{90}{votes} & 
\rotatebox{90}{frogs ($C=4$)} & 
\rotatebox{90}{frogs ($C=8$)} & 
\rotatebox{90}{frogs ($C=10$)} & 
\rotatebox{90}{isolet} & 
\rotatebox{90}{smartphone}\\
\hline
GMM & 12.46 & 84.15 & 44.74 & 49.07 & 44.22 & 25.33 & 37.22 & 20.00 & 38.99 & 18.66 & 20.71 & 19.06 & 41.72 & 69.11 & 56.82 & 27.37\\
KM & 12.17 & 71.86 & 50.67 & 47.20 & {\bf 71.09} & 66.67 & 40.56 & 60.00 & 58.55 & 87.79 & 68.49 & 7.69 & 33.80 & 54.17 & 73.78 & 63.17\\
SC & 20.70 & 84.97 & {\bf 59.66} & 43.93 & 56.12 & 55.33 & 30.28 & 62.56 & 62.76 & 32.95 & 49.46 & 23.24 & 46.44 & 72.97 & 79.72 & 52.43\\
HDB & 11.52 & 70.22 & 44.17 & 41.12 & 62.59 & 66.67 & 15.56 & 67.18 & 78.58 & 15.44 & 67.21 & 5.15 & 19.33 & 15.66 & 75.52 & {\bf 73.9}\\
MS & 12.68 & 83.88 & 50.29 & 38.32 & 52.72 & 60.67 & 44.44 & 31.79 & 73.28 & 21.89 & 21.43 & {\bf 54.42} & 39.66 & 29.15 & {\bf 85.31} & 26.30\\
SNNC & {\bf 70.77} & 66.94 & 37.09 & 45.33 & 56.80 & 66.00 & 23.89 & 73.33 & 24.44 & 78.80 & 62.72 & 7.07 & 32.66 & 25.23 & 69.06 & 60.22\\
BPC & 29.88 & 55.74 & 37.28 & 46.26 & 63.95 & 66.67 & 6.67 & {\bf 75.38} & {\bf 83.41} & {\bf 88.25} & 61.13 & 13.53 & 37.00 & 61.25 & 67.31 & 66.75\\
TCe & 18.15 & 82.24 & 36.14 & {\bf 50.93} & 35.37 & 66.67 & {\bf 46.39} & 8.72 & 82.92 & 51.61 & {\bf 70.99} & 34.59 & 27.96 & {\bf 85.35} & 76.75 & 67.91\\
TCc & 28.21 & {\bf 95.36} & 57.36 & 46.26 & 54.08 & {\bf 79.33} & 45.83 & 65.13 & 55.25 & 86.41 & 19.05 & 47.77 & 30.93 & 82.85 & 64.86 & 26.34\\
NNEC & 23.32 & 84.97 & 52.77 & 42.06 & 60.88 & 66.67 & 28.61 & 67.69 & 64.29 & 49.77 & 26.50 & 41.97 & {\bf 55.06} & 83.42 & 69.41 & 33.65\\
\hline
 & 
\rotatebox{90}{yale} & 
\rotatebox{90}{vowel} & 
\rotatebox{90}{biodeg} & 
\rotatebox{90}{ecoli} & 
\rotatebox{90}{led} & 
\rotatebox{90}{letter} & 
\rotatebox{90}{sonar} & 
\rotatebox{90}{vehicle} & 
\rotatebox{90}{wdbc} & 
\rotatebox{90}{wine} & 
\rotatebox{90}{zoo} & 
\rotatebox{90}{dna} & 
\rotatebox{90}{msplice} & 
\rotatebox{90}{musk} & 
\rotatebox{90}{pima} & 
\rotatebox{90}{spambase}\\
\hline
GMM & 37.73 & 28.08 & 59.15 & 64.88 & 51.00 & 30.07 & 53.37 & 34.87 & 78.21 & 18.01 & {\bf 86.14} & 18.90 & 51.91 & 17.20 & 16.41 & 32.91\\
KM & 68.69 & 24.04 & 62.75 & 77.08 & 55.00 & 5.71 & 52.40 & 36.88 & 91.04 & {\bf 46.72} & 39.60 & {\bf 58.8} & {\bf 55.02} & 41.83 & 52.86 & 59.86\\
SC & 61.31 & 26.26 & 35.26 & 41.37 & 54.20 & 30.39 & {\bf 65.87} & {\bf 43.62} & 62.74 & 40.28 & 81.19 & 25.25 & 51.91 & 38.44 & 64.19 & 42.64\\
HDB & {\bf 83.31} & 17.07 & 63.32 & 44.05 & 51.20 & 26.44 & 53.37 & 26.36 & 84.71 & 42.15 & 61.39 & 52.95 & 19.28 & 41.18 & 64.84 & 9.00\\
MS & 33.36 & 26.36 & 25.40 & 60.71 & 54.20 & 20.09 & 43.75 & 22.81 & 85.06 & 32.83 & 68.32 & 54.30 & 53.20 & 5.08 & 52.73 & 38.19\\
SNNC & 51.30 & 9.39 & 52.51 & 63.39 & 70.40 & 4.06 & 51.44 & 28.96 & 59.40 & 40.03 & 74.26 & 52.10 & 51.50 & {\bf 54.2} & 63.28 & 47.64\\
BPC & 76.82 & 14.14 & {\bf 66.26} & 62.50 & {\bf 76} & 28.73 & 53.37 & 37.83 & 62.74 & 42.59 & 40.59 & 52.55 & 51.91 & 26.43 & {\bf 65.1} & {\bf 61.66}\\
TCe & 78.19 & {\bf 30.71} & 38.86 & 63.69 & 48.60 & 32.09 & 30.77 & 23.88 & 83.30 & 8.13 & 75.25 & 32.05 & 14.20 & 5.84 & 15.89 & 16.28\\
TCc & 36.25 & 25.96 & 15.17 & {\bf 78.87} & 40.60 & 27.57 & 34.62 & 34.87 & {\bf 94.9} & 36.77 & 75.25 & 1.80 & 1.45 & 0.32 & 17.84 & 10.32\\
NNEC & 42.56 & 26.16 & 48.25 & 76.19 & 73.20 & {\bf 34.8} & 52.88 & 36.88 & 92.79 & 43.59 & 81.19 & 40.05 & 54.80 & 13.41 & 43.49 & 7.09\\

    \end{tabular}}
    \caption{Accuracy for all methods across all data sets}
    \label{tab:acc_all}
\end{table*}

We evaluate the performance of clustering solutions by how well they align with the true groups in the data, using the Adjusted Mutual Information~\citep[AMI]{AMI}, the Adjusted Rand Index~\citep[ARI]{hubert1985comparing}, and the clustering accuracy. The Rand Index is given by the proportion of all pairs of points which are either grouped together in both the true and estimated clusters or grouped separately in both the true and estimated clusters. The adjusted index modifies this proportion to account for the expected Rand Index under a random clustering. The adjustment to the standard mutual information between the two groupings (true and estimated clusters) applied in the AMI is similar, in that it accounts for the expected mutual information under a random clustering. The clustering accuracy is simply equal to the proportion of points falling in the same groups under the true and estimated clusters, after an optimal permutation of cluster labels.

The detailed tabulated performances from all 45 data sets and from all methods are given in Tables~\ref{tab:nmi_all},~\ref{tab:ari_all}, and~\ref{tab:acc_all}. Here we discuss summaries of their contents. To combine the results from multiple data sets, which represent clustering problems of varying difficulty level in the abstract, we standardise the results as follows. For a given performance metric (AMI, ARI or Accuracy), let $M_{i,j}$ be the value of the metric from the $j$-th clustering method on the $i$-th data set. We then define
\begin{align*}
    \mathrm{Rank}^{M}_{i,j} &:= \sum_{l=1}^L I(M_{i,l} \leq M_{i,j})\\
    M^{*}_{i,j} &:= \frac{M_{i,j} - \min_l M_{i,l}}{\max_l M_{i,l} - \min_l M_{i,l}}\\
    M^{**}_{i,j} &:= \frac{M_{i,j} - \bar{M}_{i}}{s(M)_i},
\end{align*}
where $L$ is the total number of clustering methods used (in our case 10); $I(\cdot)$ is the indicator function; and $\bar{M}_i$ and $s(M)_i$ are the mean and standard deviation of the values $\{M_{i,l}\}_{l\in [L]}$. In other words, the rank of a method on a data set is given by the number of methods which performed no better than it, i.e., the rank of the best performing method on a data set is always $L$. The quantities $\{M^*_{i,j}\}_{i,j}$ and $\{M^{**}_{i,j}\}_{i,j}$ are, respectively, the raw performances mapped to the $[0,1]$ interval and the studentised performances.
The distributions of these three standardisations of the three clustering performance metrics are shown in Figures~\ref{fig:rank},~\ref{fig:map01} and~\ref{fig:studentised}. In addition to the quartiles indicated by the box-plots, we also show the mean standardised performance values with red dots. Note that the ordering of the methods in each (sub-)figure is based on the ordering of the mean performance, and hence the ordering over methods is not always the same.\\
\\
Below we list some of the main take-aways from these results:
\begin{enumerate}
    \item The proposed NNEC achieves the highest average performance across all metrics and using all standardisations.
    \item The Torque Clustering method (specifically when using cosine distance) achieves easily the second best performance overall, and its median performance in terms of AMI exceeds that of NNEC. In fact TCc's median rank in terms of AMI is 9, meaning that it performed among the top two methods in at least half of the 48 instances, which is quite exceptional over such a large collection of data sets.
    \item Torque Clustering performs comparatively poorly in terms clustering accuracy, suggesting it tends to over-cluster to a greater extent than NNEC, which can be reasoned by the fact that clustering accuracy is far more heavily penalising of over-clustering than either of AMI and ARI.
    \item Although not evident from these figures, the Border Peeling Clustering approach (BPC) has strongly bimodal performance, achieving quite high performance in a large proportion of cases but also failing to identify any clusters in more than a third of the data sets (16 out of 45). We followed the authors' recommendations in setting the number of neighbours equal to 20, but it is likely that a setting which depends on $n$ is more appropriate in general and may have changed the overall layout of results considerably. It is very important to note, however, that even if we restrict attention to those data sets on which BPC did \textit{not fail} to identify clusters (artificially inflating its realistic performance) the proposed approach still achieves higher overall performance using all metrics and all standardisations. This highlights the importance of extensive experimental comparisons, where likely had the authors considered a greater variety of data sets they would have identified this limitation in their recommendation.
    \item Although Spectral Clustering (SC) only achieved the highest performance in a small number of instances, it was among the better performing methods on the majority and is arguably the third best performing method overall. A possible contender to this is Mean Shift (MS), however the results of MS are arguably less consistent.
    \item Despite its simplicity, $K$-means (KM) consistently produces solutions which are at least reasonably good, and it is among the better performing methods overall. In addition, the high clustering accuracy of KM, as we have applied it, suggests the silhouette index has performed very well in appropriately selecting the number of clusters, where we discussed previously the heavy penalisation by this metric of over-clustering.
\end{enumerate}

\section{Conclusions}\label{sec:conclusions}

In this paper we introduced a simple and intuitive clustering method based on the natural principle that points should be grouped along with a high proportion of their nearest neighbours. Recognising that the extent to which this is achievable will be heavily data dependent, we introduced a natural criterion which can be used to automatically determine appropriate tuning parameters for the method.

In an extensive set of experiments, over 45 data sets commonly used in the clustering literature, the proposed approach achieved consistently high performance as measured by Adjusted Rand Index, Adjusted Mutual Information and clustering accuracy, and in comparison with relevant benchmarks both new and old.





\bibliographystyle{ieeetr}


\end{multicols}

%






\end{document}